\documentclass[11pt]{article}

\usepackage[preprint]{acl}

\usepackage{times}
\usepackage{latexsym}

\usepackage[T1]{fontenc}

\usepackage[utf8]{inputenc}

\usepackage{microtype}
\usepackage{xurl}

\usepackage{inconsolata}

\usepackage{graphicx}
\usepackage[most]{tcolorbox}
\usepackage{fancyvrb}
\usepackage{fvextra}

\newcounter{codecalgorithm}
\newcommand{\methodfull}{Collaboration Policy Tree}
\newcommand{\method}{Co-\texorpdfstring{$\pi$}{pi}-tree}

\newtcolorbox{policybox}[1]{
  enhanced,
  breakable,
  colback=gray!3,
  colframe=black!60,
  boxrule=0.5pt,
  arc=1pt,
  left=3pt,
  right=3pt,
  top=3pt,
  bottom=3pt,
  title={#1},
  fonttitle=\bfseries,
  coltitle=black,
  colbacktitle=gray!15
}

\DefineVerbatimEnvironment{PolicyVerb}{Verbatim}{
  fontsize=\tiny,
  breaklines=true,
  breakanywhere=true,
  breaksymbolleft={},
  breaksymbolright={}
}
\DefineVerbatimEnvironment{PolicyVerbRed}{Verbatim}{
  fontsize=\tiny,
  breaklines=true,
  breakanywhere=true,
  breaksymbolleft={},
  breaksymbolright={},
  formatcom=\color{red!75!black}
}
\DefineVerbatimEnvironment{PolicyVerbGreen}{Verbatim}{
  fontsize=\tiny,
  breaklines=true,
  breakanywhere=true,
  breaksymbolleft={},
  breaksymbolright={},
  formatcom=\color{green!45!black}
}

\title{Distilling LLM Reasoning into an Interpretable Policy Tree \\for Human-AI Collaboration}

\author{
  Beiwen Zhang\thanks{Equal contribution.},
  Yongheng Liang\footnotemark[1],
  Guowei Zou,
  Haitao Wang, and
  Hejun Wu\thanks{Corresponding author.} \\
  Sun Yat-sen University \\
  {\small \texttt{zhangbw39@mails.sysu.edu.cn}, \texttt{liangyh38@mail2.sysu.edu.cn}, \texttt{zougw@mail2.sysu.edu.cn}} \\
  {\small \texttt{wanght76@mail.sysu.edu.cn}, \texttt{wuhejun@mail.sysu.edu.cn}}
}

\begin{document}
\frenchspacing
\maketitle
\begin{abstract}
Constructing efficient and reliable policies to assist humans is indispensable for human-AI collaboration. Existing methods mainly follow two lines of work. The major parts of prior work rely on multi-agent reinforcement learning (MARL) to learn black-box policies. This limits the interpretability and raises safety concerns. Recent methods query large language models (LLMs) at each decision step, causing slow responses and high inference costs. We propose \methodfull{} (\method{}), a closed-loop method that learns an executable policy tree consisting of a partner-behavior prediction tree and an agent-action selection tree. \method{} constructs a policy through distilling LLM reasoning into policy tree code. \method{} then evaluates the policy through partner interaction and obtains feedback, and uses natural language to summarize interaction feedback to improve problematic branches. Experiments in Overcooked-AI show that \method{} improves average reward by 35.4\% over the baseline average, while reducing the number of LLM queries by 77.7\% and test-time latency by 97.1\%. Project page: https://beiwenzhang.github.io/Co-pi-tree/.
\end{abstract}

\section{Introduction}
Constructing efficient and reliable policies that enable AI agents to better assist human partners is essential for human-AI collaboration, especially in domains such as healthcare, autonomous driving, and assistive robotics \citep{Topol2019HighPM,Paden2016Survey}. A core
challenge in human-AI collaboration is zero-shot coordination (ZSC), where
agents must cooperate effectively with previously unseen partners
\citep{Carroll2019OnTU,FirstZsc}. Most existing ZSC methods rely on
multi-agent reinforcement learning (MARL), typically training an agent to
cooperate with simulated partners that cover different behaviors and strategies
\citep{FCP,Zhao2021MEP,COLE}. Their effectiveness therefore depends on partner diversity, while their black-box policies limit interpretability and may lead to unpredictable or unsafe actions in human
coordination \citep{Endsley2023SupportingHT,ahin2024UnlockingTB,Anne2024HarnessingLF}.

Since human collaboration relies heavily on language for establishing common ground, sharing intentions, and coordinating joint actions \citep{Clark1991Grounding}, natural language provides an effective medium for human-AI collaboration. With strong language understanding, reasoning, and planning abilities \citep{Du2025ASO,Shinn2023ReflexionLA,Liu2023AgentBenchEL}, large language models (LLMs) offer a promising tool for supporting such language-mediated collaboration. Recent studies have begun to operationalize this idea by converting observations and partner behaviors into textual contexts, querying LLMs to generate coordination plans, and translating the responses into executable actions \citep{Zhang2023ProAgentBP,Liu2023LLMPoweredHL,Sun2025CollabOvercookedBA}.

\begin{figure*}[t]
\centering
\includegraphics[width=1.0\textwidth]{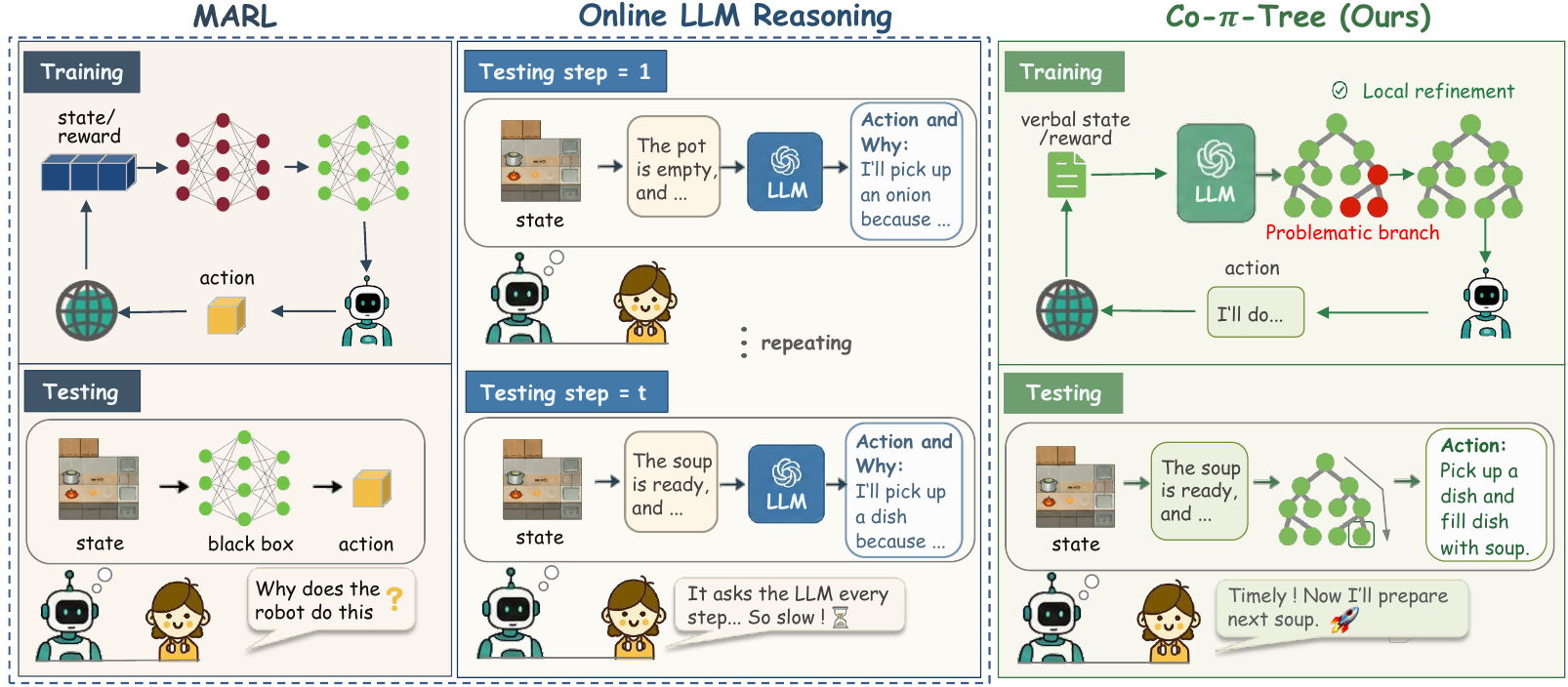}
\caption{Left: MARL methods update all policy parameters during training and directly use the learned black-box policy to collaborate with humans. Middle: online LLM agents require no training but query the LLM before every decision. Right: \method{} uses the LLM to locate and revise problematic branches, and directly executes the refined policy tree with humans.}
\label{fig:motivation}
\end{figure*}

However, current LLM-based ZSC collaboration methods still face an efficiency bottleneck in real-time coordination. They typically require an LLM query before each action, causing delayed responses and high inference costs \citep{Zhang2023ProAgentBP,Liu2023LLMPoweredHL,Zhang2024TowardsEL}. In contrast, MARL-based methods support low-latency execution but learn black-box policies with limited interpretability. We therefore ask: \textit{Can we obtain executable policies for human-AI collaboration that preserve the language-based coordination reasoning of LLMs while requiring only limited LLM queries and low test-time latency?}

Since tree structures offer fast and interpretable execution by selecting actions through explicit condition branches \citep{Rudin2018StopEB,Xiong2024GPTreeTE}, we propose \methodfull{} (\method{}), which distills LLM-based coordination reasoning into an executable policy tree. \method{} contains two components: a partner-behavior prediction tree that predicts how the partner may act, and an agent-action selection tree that adapts to the predicted partner behavior to choose cooperative actions. \method{} training consists of three stages: (1) the policy
construction stage distills LLM reasoning into executable policy tree code; (2)
the environment grounding stage evaluates the policy through partner interaction and
obtains feedback; and (3) the policy refinement stage uses natural language to summarize
interaction feedback and improve problematic branches. The resulting policy tree can then be directly executed without querying the LLM at each decision step.

Our work makes three contributions. (1) We introduce a policy-tree structure for human-AI collaboration, enabling efficient and interpretable execution. (2) We design a closed-loop algorithm that distills LLM reasoning into a policy tree and refines problematic branches through natural-language feedback. (3) We validate
\method{} in Overcooked-AI \citep{Carroll2019OnTU} with AI and human partners, showing that it improves average reward by 35.4\% over the baseline average, while reducing the number of LLM queries by 77.7\% and test-time latency by 97.1\%.

\section{Related Work}

\noindent\textbf{ZSC in Human-AI Collaboration.}
To assist different humans effectively, collaborative policies need to generalize to previously unseen partners at test time \citep{Carroll2019OnTU}, which is the ZSC problem in human-AI collaboration \citep{FirstZsc,Gessler2025OvercookedV2RO}. Existing MARL-based ZSC methods typically first train a diverse set of simulated partner policies, and then train an agent to cooperate with these partners to improve generalization to unseen human partners \citep{FCP, Zhao2021MEP, COLE, Jha2025CrossenvironmentCE}. However, their performance depends heavily on the diversity and coverage of the simulated partners, while the learned black-box policies limit interpretability and human oversight \citep{Endsley2023SupportingHT, ahin2024UnlockingTB,Anne2024HarnessingLF}.

\begin{figure*}[t]
\centering
\includegraphics[width=1.0\textwidth]{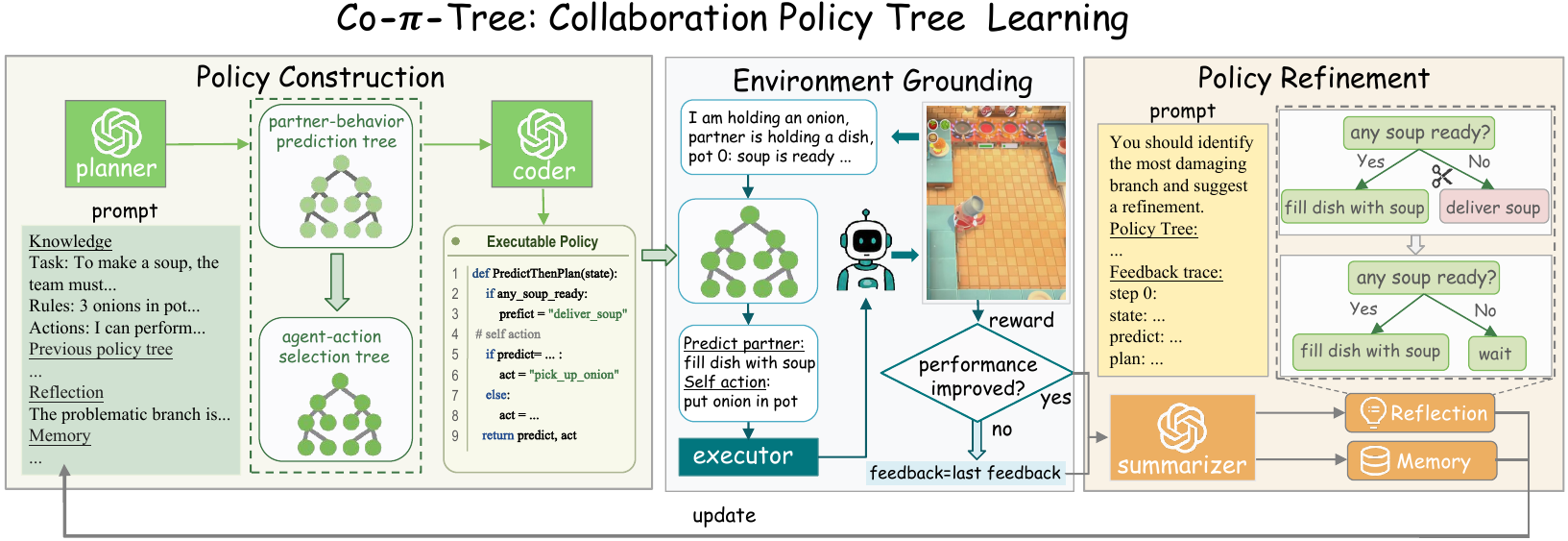}
\caption{Overview of the \method{} pipeline.}
\label{fig:framework}
\end{figure*}

\noindent\textbf{LLM Reasoning and Coordination.}
LLMs have shown strong capabilities in multi-step reasoning and planning across a wide range of tasks \citep{Wei2022ChainOT, Zhou2024SelfDiscoverLL, Yao2022ReActSR}. Beyond direct online generation, some studies convert LLM outputs into reusable decision forms, such as executable programs or symbolic logic, and refine them with feedback \citep{Gallego2026CooperationAE, Hu2024TeachingLM, Xiong2024GPTreeTE}. These studies suggest that LLM reasoning can be used not only for one-time response generation, but also for constructing decision structures that can be reused and improved.

Recent studies have also explored LLM-based reasoning for collaborative decision making. ProAgent verbalizes observations and partner behaviors to infer intent and replan actions \citep{Zhang2023ProAgentBP}, HLA uses natural-language communication for command interpretation and macro-action generation \citep{Liu2023LLMPoweredHL}, and CausalPlan augments LLM action selection with causal scores over candidate actions \citep{Nguyen2026CausalPlan}. These methods demonstrate the potential of language-based reasoning for coordination. However, they often rely on LLM-driven decisions during interaction, requiring repeated inference and introducing latency and inference cost \citep{Zhang2024TowardsEL, Parashar2025InferenceTimeCF, Xiao2025LIMOProRR, Pan2025SpecReasonFA}.

\section{Methodology}
To address the opacity of MARL policies and the high test-time cost of online
LLM agents, we propose \method{}, a policy learning algorithm that
distills LLM reasoning into an executable policy tree.
Figure~\ref{fig:framework} illustrates the overall pipeline, which learns an
interpretable policy through three stages. In the policy construction stage,
the planner and coder generate a candidate policy tree and
translate it into executable code. In the environment grounding stage, the
executor runs the candidate policy through partner interaction and obtains natural-language
feedback. The resulting reward determines whether the
candidate policy is accepted. In the policy refinement stage, the summarizer
uses the feedback to identify problematic branches and produce a reflection for
the next iteration.

\subsection{Dec-POMDP}
We formulate the human-AI collaboration problem as a Decentralized Partially
Observable Markov Decision Process (Dec-POMDP),
\begin{equation}
    \mathcal{G}=(\mathcal{I}, \mathcal{S}, \mathcal{A}, \Omega, P, Z, R, \gamma),
\end{equation}
where $\mathcal{I}$ is the agent set, $\mathcal{S}$ is the state space,
$\mathcal{A}$ is the joint action space, $\Omega$ is the joint observation
space, $P$ and $Z$ are the transition and observation functions, respectively,
$R$ is the shared team reward, and $\gamma$ is the discount factor. 

The problem objective is to learn a policy $\pi$ for the controlled
agent that can be executed with previously unseen partners. We optimize $\pi$ to maximize
episodic team reward over horizon $H$:
\begin{equation}
    J(\pi)=\mathrm{E}_{\tau \sim \pi}\!\left[\sum_{t=0}^{H-1}\gamma^t r_t\right].
\end{equation}

\subsection{Policy Construction Stage}
\label{sec:policy-construction}
This stage converts LLM reasoning into a structured and executable policy. It first uses the planner to generate a candidate
two-component policy tree $T$, and then uses the coder to translate $T$
into executable code $c$. 

\noindent\textbf{Planner: Policy Tree Construction.}
Let $\mathcal{M}$ denote the LLM and $\Vert$ denote prompt concatenation.
At the $k$-th iteration, the planner takes as input two specialized prompts
$p_{\mathrm{pred}}$ and $p_{\mathrm{act}}$, memory $\mathcal{B}_{<k}$, which
stores concise summaries of previously accepted policies, the previous accepted
policy tree $T^\star_{k-1}$, and the corresponding reflection $\rho_{k-1}$.

The planner first uses $p_{\mathrm{pred}}$ to generate $T^{\mathrm{pred}}_k$, which predicts the partner's likely
behavior. It then uses $T^{\mathrm{pred}}_k$ to generate $T^{\mathrm{act}}_k$, which selects the controlled agent's
action conditioned on the predicted partner behavior. Together, these
two trees form the candidate policy tree $T_k$:
\begin{equation}
\begin{array}{@{}l@{\;}c@{\;}l@{}}
    T^{\mathrm{pred}}_k
    & = & \mathcal{M}\bigl(p_{\mathrm{pred}} \Vert \mathcal{B}_{<k}
       \Vert T^\star_{k-1} \Vert \rho_{k-1}\bigr), \\
    T^{\mathrm{act}}_k
    & = & \mathcal{M}\bigl(p_{\mathrm{act}} \Vert \mathcal{B}_{<k}
       \Vert T^\star_{k-1} \Vert \rho_{k-1}
       \Vert T^{\mathrm{pred}}_k\bigr), \\
    T_k
    & = & (T^{\mathrm{pred}}_k, T^{\mathrm{act}}_k).
\end{array}
\end{equation}

The generated trees operate on a structured policy input
$x_t=\phi(s_t,h_t)$, where $s_t$ is the current environment state and $h_t$
records recent agent behavior. The mapping $\phi$ converts raw observations
into a readable context that exposes task progress, partner intent, and
coordination needs. During execution, the
policy first predicts the partner's behavior and then selects the controlled
agent's action:
\begin{equation}
    \hat{z}_t = T^{\mathrm{pred}}_k(x_t), \qquad
    a_t = T^{\mathrm{act}}_k(x_t,\hat{z}_t).
\end{equation}
Here, $\hat{z}_t$ denotes the predicted partner behavior used by the
agent-action selection tree.
We use a tree structure for $\pi$ because it is easy for humans to inspect and revise.
Each branch is an explicit \texttt{if/elif} condition-action rule, exposing why
a partner behavior is predicted and why the controlled agent selects an action in a
given scene. This locality supports branch-level diagnosis and revision, which
is preferable to opaque neural policies when decisions need to be audited
\citep{Rudin2018StopEB}.

\noindent\textbf{Coder: Executable Code Generation.}
The coder takes the textual policy tree $T_k$ as input and feeds it, together
with the code-generation prompt $p_{\mathrm{code}}$, into the LLM:
\begin{equation}
    c_k = \mathcal{M}(p_{\mathrm{code}} \Vert T_k).
\end{equation}
The output is an executable Python function $c_k$ that implements the same
partner-behavior prediction and agent-action selection logic:
\begin{equation}
    c_k(x_t) = (\hat{z}_t, a_t).
\end{equation}
The coder preserves the policy tree logic with explicit \texttt{if/elif}
branches, while grounding to executable environment actions and feasibility
checks are handled by the executor.

\noindent\textbf{Prompt and Output Design.}
$p_{\mathrm{pred}}$ and $p_{\mathrm{act}}$ specialize the planner into two
connected generation tasks. $p_{\mathrm{pred}}$ asks the LLM to generate a tree for predicting the partner's likely behavior. It includes a knowledge library
(task objective, task rules, and available actions), the required
tree structure, and the input-output format. This enables the LLM to express
partner prediction as explicit decision branches.
$p_{\mathrm{act}}$ asks the LLM to generate a tree for selecting the controlled
agent's action. Its prompt composition is similar to
$p_{\mathrm{pred}}$, but it additionally requires the action tree to condition
on the predicted partner-behavior tree, so the selected action can be
complementary to the predicted partner behavior. We also include a small set of demonstration scenes to ground action
semantics and clarify how scene conditions and partner behavior should affect
branch construction \citep{Brown2020LanguageMA,Dong2023ASF}.

Both the planner and coder are constrained to follow prescribed output
structures and code interfaces
\citep{openai2024structuredoutputs,Wang2025SLOTST}. These constraints turn LLM
generations into machine-readable policy tree descriptions rather than
free-form text, and allow the coder to translate the tree into executable
functions. Detailed prompt
schemas, demonstration format, and code interface constraints are provided in
Appendix~\ref{app:prompts}.

\subsection{Environment Grounding Stage}
This stage connects the executable code with environment interaction by
grounding tensor states into textual states, grounding tree output actions into
executable environment actions, and returning feedback for policy refinement.

\noindent\textbf{Tensor States to Textual States.}
\method{} first grounds tensor states into textual states that the LLM can read and uses task semantic fields to describe the current scene. Tensor states in
RL environments encode environment variables such as agent locations,
orientations, object states, and layout configurations, while textual states
make task progress, object status, and partner behavior explicit for LLM
reasoning. For execution, these textual states are then
converted into an input dictionary with a fixed format, which serves as the
input to the policy tree code.
The full grounding schema is provided in
Appendix~\ref{app:environment-grounding}.

\noindent\textbf{Output Actions to Executable Actions.}
Tree output actions are mapped to executable environment actions by an
executor. In Overcooked-AI, the executor maps an action, such as picking up an
onion, to executable environment actions such as movement and interaction
commands. The executor selects a feasible destination and the next executable
environment action, and logs failures in the execution trace when the action is
not executable. This separation allows the LLM to focus on action reasoning,
while the executor handles navigation, reachability, and other environment
constraints.

\noindent\textbf{Executor and Feedback.}
The executor evaluates a candidate policy by running the code in the
environment. Across an episode, it returns feedback
$\mathcal{F}_k=(S_k,\tau_k)$, where $S_k$ is the evaluation score and
$\tau_k$ is the grounded execution
trace. The trace records the scene state, whether the selected action was
executable, and whether the predicted partner behavior is correct. The realized
partner behavior is derived from the action labels recorded by the
environment after each interaction.
The score $S_k$ determines whether the candidate policy is accepted. If the
candidate improves or matches the best accepted score $S^\star_{k-1}$,
\method{} accepts the candidate policy tree; otherwise, it reverts to the
previous accepted policy tree. The accepted policy and its feedback are
updated as
\begin{equation}
    (T^\star_k,\mathcal{F}^\star_k) =
    \left\{
    \begin{array}{ll}
        (T_k,\mathcal{F}_k), & \mathrm{if}\ S_k \ge S^\star_{k-1},\\
        (T^\star_{k-1},\mathcal{F}^\star_{k-1}), & \mathrm{otherwise}.
    \end{array}
    \right.
\end{equation}

\subsection{Policy Refinement Stage}

This stage converts feedback into language-level updates for the next
policy construction stage. Given the accepted policy tree $T^\star_k$ and its
feedback $\mathcal{F}^\star_k$, \method{} produces a policy summary $m_k$ for
memory and a reflection $\rho_k$ for refining problematic branches in the next
planner call.

\noindent\textbf{Summarizer: Reflection from Feedback.}
The input to the summarizer is the accepted policy tree $T^\star_k$ and its
feedback $\mathcal{F}^\star_k=(S^\star_k,\tau^\star_k)$. The execution trace
$\tau^\star_k$ contains compact language-based scene descriptions, which
allow the summarizer to analyze feedback at the scene level; the trace format
and example entries are given in Appendix~\ref{app:environment-grounding}.
Following the principle of verbal reinforcement learning
\citep{Shinn2023ReflexionLA}, the summarizer feeds this information, together
with a summarization prompt $p_{\mathrm{sum}}$, into the LLM to generate
information for updating the partner-behavior prediction tree and the
self-action selection tree:
\begin{equation}
    g_k = (m_k,\rho_k)
    = \mathcal{M}(p_{\mathrm{sum}} \Vert T^\star_k \Vert \mathcal{F}^\star_k),
\end{equation}
where $m_k$ is a compact policy tree summary and $\rho_k$ is a reflection for
policy refinement. 

This design converts sparse execution feedback into
structured language supervision. The reflection $\rho_k$ identifies a damaging
or inefficient branch, explains the failure pattern observed in the feedback,
and proposes a localized reflection for the next planner call. This
branch-level reflection addresses credit assignment, which refers to
identifying which decision branch is responsible for an episode failure. This
is challenging because sparse team reward does not directly reveal whether the
failure comes from a specific local decision, such as an incorrect
partner-behavior prediction or premature dish handling in Overcooked-AI
\citep{Zhang2026FromRT}. Because each decision corresponds to an explicit
branch, the summarizer can assign episode failures to local branches of the
policy tree rather than returning only a global critique. The planner then
preserves unchanged branches and refines only the exposed branch, reducing
uncontrolled changes to unrelated branches.

\noindent\textbf{Memory.}
Past accepted policies can inform future planning \citep{Madaan2023SelfRefineIR},
so \method{} maintains a memory pool $\mathcal{B}$ of accepted
policy tree summaries $m_k$. This memory enables reuse of high-quality
policies and helps the planner prioritize effective behaviors. To avoid
placing all historical policies and feedback into the LLM context, which can
increase inference cost and dilute key information in long-context prompting
\citep{Jiang2023LongLLMLinguaAA}, \method{} stores concise summaries and
instructs the planner to prefer higher-scoring policies.

Algorithm~\ref{alg:codec-tree} summarizes the \method{} policy-learning
loop. Additional parameter details are provided in
Appendix~\ref{app:algorithm}.

\refstepcounter{codecalgorithm}\label{alg:codec-tree}
\begin{figure}[t]
\small
\fbox{
\begin{minipage}{0.82\linewidth}
\textbf{Algorithm~\thecodecalgorithm: \method{} Policy Learning}
\begin{tabbing}
\quad\=\quad\=\quad\=\kill
\textbf{Input:} LLM $\mathcal{M}$, number of iterations $K$\\
\textbf{Initialize:} $\mathcal{B}\leftarrow\emptyset$,
$T^\star\leftarrow\emptyset$, $c^\star\leftarrow\emptyset$,
$\mathcal{F}^\star\leftarrow\emptyset$,\\ $S^\star\leftarrow-\infty$,
$\rho\leftarrow\emptyset$, $u\leftarrow 0$\\
\textbf{for} $k=1,\ldots,K$ \textbf{do}\\
\> $\vartheta_k \leftarrow \min(\vartheta_{\max}, \vartheta_{\mathrm{base}}(1+\log(1+u)))$\\
\> $T^{\mathrm{pred}}_k \leftarrow \mathrm{plannerInf}(\mathcal{B},T^\star,\rho,\vartheta_k)$\\
\> $T^{\mathrm{act}}_k \leftarrow \mathrm{plannerAct}(\mathcal{B},T^\star,\rho,$\\
\>\> $T^{\mathrm{pred}}_k,\vartheta_k)$\\
\> $T_k \leftarrow (T^{\mathrm{pred}}_k,T^{\mathrm{act}}_k)$\\
\> $c_k \leftarrow \mathrm{coder}(T_k,\vartheta_k)$\\
\> $\mathcal{F}_k \leftarrow \mathrm{executor}(c_k)$\\
\> $S_k \leftarrow \mathrm{score}(\mathcal{F}_k)$\\
\> $S_{\mathrm{old}} \leftarrow S^\star$\\
\> \textbf{if} $S_k \ge S^\star$ \textbf{then}\\
\>\> $(m_k,\rho_k) \leftarrow \mathrm{summarizer}(T_k,\mathcal{F}_k)$\\
\>\> $T^\star \leftarrow T_k$\\
\>\> $c^\star \leftarrow c_k$\\
\>\> $\mathcal{F}^\star \leftarrow \mathcal{F}_k$\\
\>\> $S^\star \leftarrow S_k$\\
\>\> $\mathcal{B}\leftarrow\mathcal{B}\cup\{m_k\}$\\
\>\> $\rho\leftarrow\rho_k$\\
\>\> \textbf{if} $S_k > S_{\mathrm{old}}$ \textbf{then} $u\leftarrow 0$
\textbf{ else } $u\leftarrow u+1$\\
\> \textbf{else}\\
\>\> $(\_,\rho^\star) \leftarrow \mathrm{summarizer}(T^\star,\mathcal{F}^\star)$\\
\>\> $\rho\leftarrow\rho^\star$\\
\>\> $u\leftarrow u+1$\\
\> \textbf{end if}\\
\textbf{end for}\\
\textbf{return} executable policy $c^\star$ and interpretable policy tree $T^\star$
\end{tabbing}
\end{minipage}
}
\end{figure}

\section{Experiments}
\label{sec:experiments}
\providecommand{\stdev}[1]{{\scriptsize $\pm$ #1}}

\begin{table*}[t]
\centering
\small
\renewcommand{\arraystretch}{1.3}
\setlength{\tabcolsep}{2.5pt}
\begin{tabular}{lccccccccc}
\hline
 & \multicolumn{6}{c}{AI Baseline} &
\multicolumn{2}{c}{LLM Baseline} & \multicolumn{1}{c}{Ours} \\
\cline{2-10}
 & SP & PBT & FCP & MEP & COLE & BC & ProAgent & CausalPlan & \method{} \\
\hline
Cramped Rm. &
155.0\stdev{43} & 161.3\stdev{45} & \underline{175.8}\stdev{30} &
161.3\stdev{42} & 153.8\stdev{29} & 131.3\stdev{37} &
171.0\stdev{26} & 174.5\stdev{20} & \textbf{182.3}\stdev{18} \\
 &
157.5\stdev{32} & 158.8\stdev{53} & \underline{170.3}\stdev{33} &
168.8\stdev{35} & 153.8\stdev{37} & 130.0\stdev{37} &
168.3\stdev{18} & 170.2\stdev{16} & \textbf{180.1}\stdev{16} \\
\hline
Coord. Ring &
103.8\stdev{49} & 118.8\stdev{39} & 136.3\stdev{28} &
152.0\stdev{21} & 150.3\stdev{32} & 96.3\stdev{37} &
151.0\stdev{34} & \underline{155.7}\stdev{24} & \textbf{165.9}\stdev{28} \\
 &
127.5\stdev{49} & 127.5\stdev{37} & 137.5\stdev{24} &
140.0\stdev{39} & 144.0\stdev{31} & 96.3\stdev{41} &
143.3\stdev{28} & \underline{152.0}\stdev{26} & \textbf{162.1}\stdev{24} \\
\hline
CT. Circuit &
30.0\stdev{33} & 43.8\stdev{49} & 42.5\stdev{45} &
53.8\stdev{35} & 85.0\stdev{30} & 47.5\stdev{35} &
108.3\stdev{24} & \underline{108.8}\stdev{21} & \textbf{117.8}\stdev{15} \\
 &
36.3\stdev{26} & 35.0\stdev{39} & 37.5\stdev{45} &
65.0\stdev{41} & 86.3\stdev{34} & 40.0\stdev{35} &
\underline{106.0}\stdev{19} & 104.9\stdev{21} & \textbf{116.0}\stdev{17} \\
\hline
Asymm. Adv. &
151.3\stdev{60} & 147.5\stdev{73} & 141.3\stdev{68} &
116.3\stdev{74} & 187.5\stdev{46} & 150.0\stdev{56} &
\underline{256.7}\stdev{31} & 250.4\stdev{25} & \textbf{274.6}\stdev{31} \\
 &
190.0\stdev{32} & 136.3\stdev{76} & 170.0\stdev{46} &
177.5\stdev{59} & 150.0\stdev{70} & 82.5\stdev{75} &
228.3\stdev{15} & \underline{228.6}\stdev{20} & \textbf{239.1}\stdev{18} \\
\hline
Forced Coord. &
12.5\stdev{17} & 21.3\stdev{22} & 56.3\stdev{42} &
23.8\stdev{25} & 40.0\stdev{32} & 40.0\stdev{23} &
56.7\stdev{22} & \textbf{64.3}\stdev{25} & \underline{62.7}\stdev{31} \\
 &
28.8\stdev{25} & \textbf{61.3}\stdev{42} & 18.8\stdev{26} &
35.0\stdev{32} & \underline{41.3}\stdev{24} & 21.3\stdev{22} &
33.3\stdev{34} & 34.5\stdev{28} & 35.6\stdev{22} \\
\hline
\end{tabular}
\caption{ZSC with AI partners. Each entry reports mean team
reward $\pm$ std. For each layout, the two rows correspond to assigning the
evaluated policy to Player 0 and Player 1, respectively. Bold and underline
denote the best and second-best result in each layout-role row.}
\label{tab:ai-main}
\end{table*}

We evaluate \method{} in the Overcooked-AI benchmark
\citep{Carroll2019OnTU}, a standard testbed for ZSC and human-AI collaboration.
Our experiments are designed to answer three questions:
\textbf{Q1.} How effectively does \method{} collaborate with both unseen AI
partners and real human partners compared with existing ZSC methods?
\textbf{Q2.} How do the main components of \method{}, including partner
prediction and iterative refinement, affect collaborative performance?
\textbf{Q3.} How scalable is the learned policy tree when transferred across
layouts, and how interpretable is it when inspected through its explicit
decision branches?
\subsection{Experimental Setup}

\noindent\textbf{Environments.}
We use five layouts from Overcooked-AI: \textit{cramped room},
\textit{coordination ring}, \textit{counter circuit},
\textit{asymmetric advantages}, and \textit{forced coordination}, abbreviated as
\textit{Cramped Rm.}, \textit{Coord. Ring}, \textit{CT. Circuit},
\textit{Asymm. Adv.}, and \textit{Forced Coord.}, respectively.
A detailed description
of each layout is provided in the appendix.
Each episode
lasts 400 environment steps, and performance is measured by the team reward,
which is determined by the number of delivered soups.

\noindent\textbf{Baselines.}
We compare with two groups of ZSC methods. The first group includes
MARL policies and imitation policies commonly used in ZSC: Self-Play (SP),
Population-Based Training (PBT), Fictitious Co-Play (FCP) \citep{FCP},
Maximum Entropy Population-Based Training (MEP) \citep{Zhao2021MEP},
Cooperative Open-ended Learning (COLE) \citep{COLE}, and a human proxy trained
by behavior cloning (BC) \citep{Carroll2019OnTU}. The second group includes two LLM-based methods: ProAgent \citep{Zhang2023ProAgentBP} and CausalPlan \citep{Nguyen2026CausalPlan}. Additional details on baseline selection are provided in Appendix~\ref{app:baselines}.

\noindent\textbf{\method{} Variants.}
To understand the roles of partner-behavior prediction and
partner-conditioned action selection, we design two ablation variants.
\method{} is the full two-component method introduced in
Section~\ref{sec:policy-construction}.
\method{}-PI asks the LLM to generate both trees, but the agent-action selection tree does
not explicitly use the predicted partner behavior. In this variant, the
partner-behavior prediction tree serves only as an intermediate reasoning step
for generating the agent-action selection tree, similar to chain-of-thought
prompting.
\method{}-w/o P removes partner-behavior prediction and directly asks the LLM
to generate an agent-action selection tree independent of partner behavior.
Additional details are provided in Appendix~\ref{app:algorithm}.

\begin{figure*}[t]
\centering
\includegraphics[width=\textwidth]{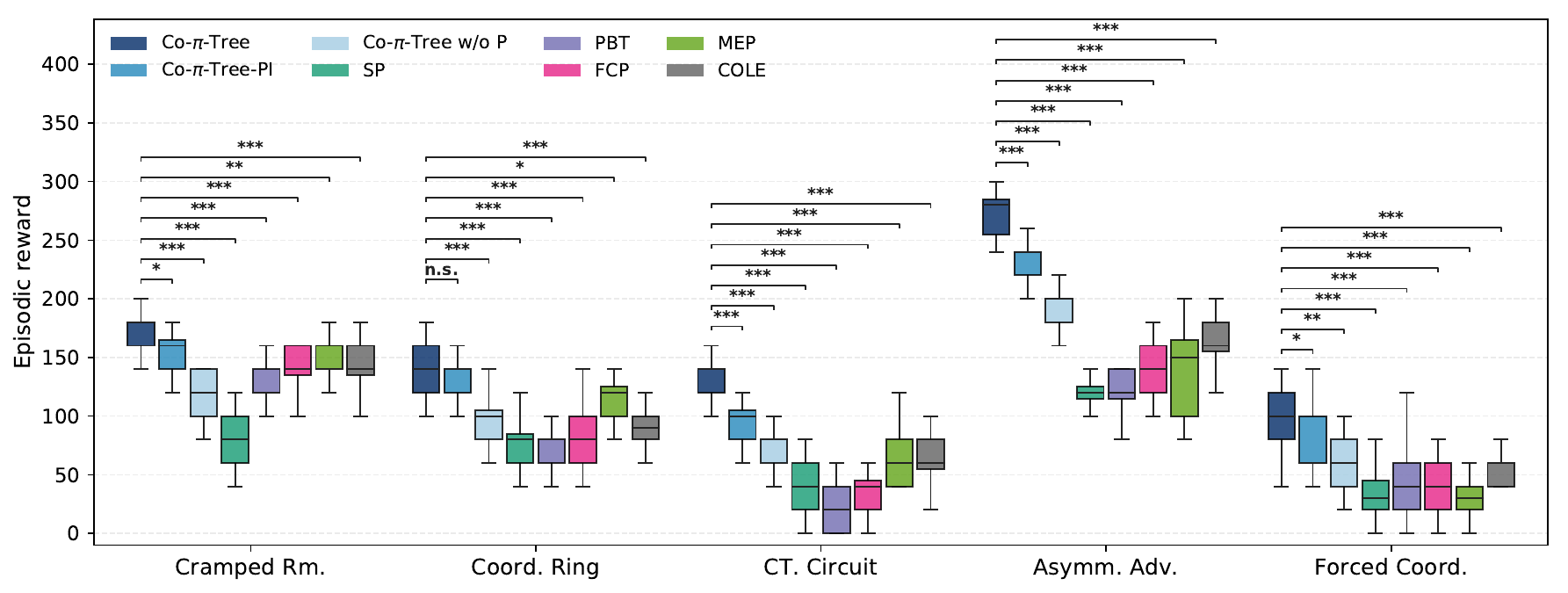}
\caption{Box plots of human-agent collaboration rewards by layout. Each box
shows per-volunteer average rewards for one method, averaging the two
player-role assignments. Significance markers compare each method with
\method{} within the same layout using Holm-corrected two-sided Mann--Whitney U
tests; n.s. denotes not significant, and *, **, *** denote $p<0.05$, $p<0.01$,
and $p<0.001$.}
\label{fig:human-box}
\end{figure*}

\noindent\textbf{Evaluation Protocol.}
We follow the standard ZSC evaluation protocol. For each evaluated method, the
agent controlled by this method is paired with each partner agent from
\{SP, PBT, FCP, MEP, COLE, BC\}, and we report the average team reward across all
pairings. We evaluate both role assignments, where the evaluated method controls
Agent 0 and Agent 1, respectively. We also report NQ as the total number of LLM
queries and latency as the average test-time decision time. Further details on
the evaluation protocol are provided in Appendix~\ref{app:ai-protocol}.

\subsection{Collaborating with Other AI Partners}
To answer Q1 in the standard ZSC setting with unseen AI partners,
Table~\ref{tab:ai-main} reports the main results against the MARL, BC, and
LLM-based baselines. \method{} achieves strong performance across all layouts,
outperforming all baselines in 8 of the 10 layout-role settings. Averaged over
all layout-role settings, \method{} improves average reward by 35.4\% over
the baseline average. The gains are especially clear in
CT. Circuit and Asymm. Adv., where successful collaboration
requires stable role specialization.
The main exception is Forced Coord., where \method{} is competitive as Player 0
but weaker as Player 1. We provide a brief discussion in
Appendix~\ref{app:forced-coordination}.

\begin{figure}[t]
\centering
\includegraphics[width=0.47\textwidth]{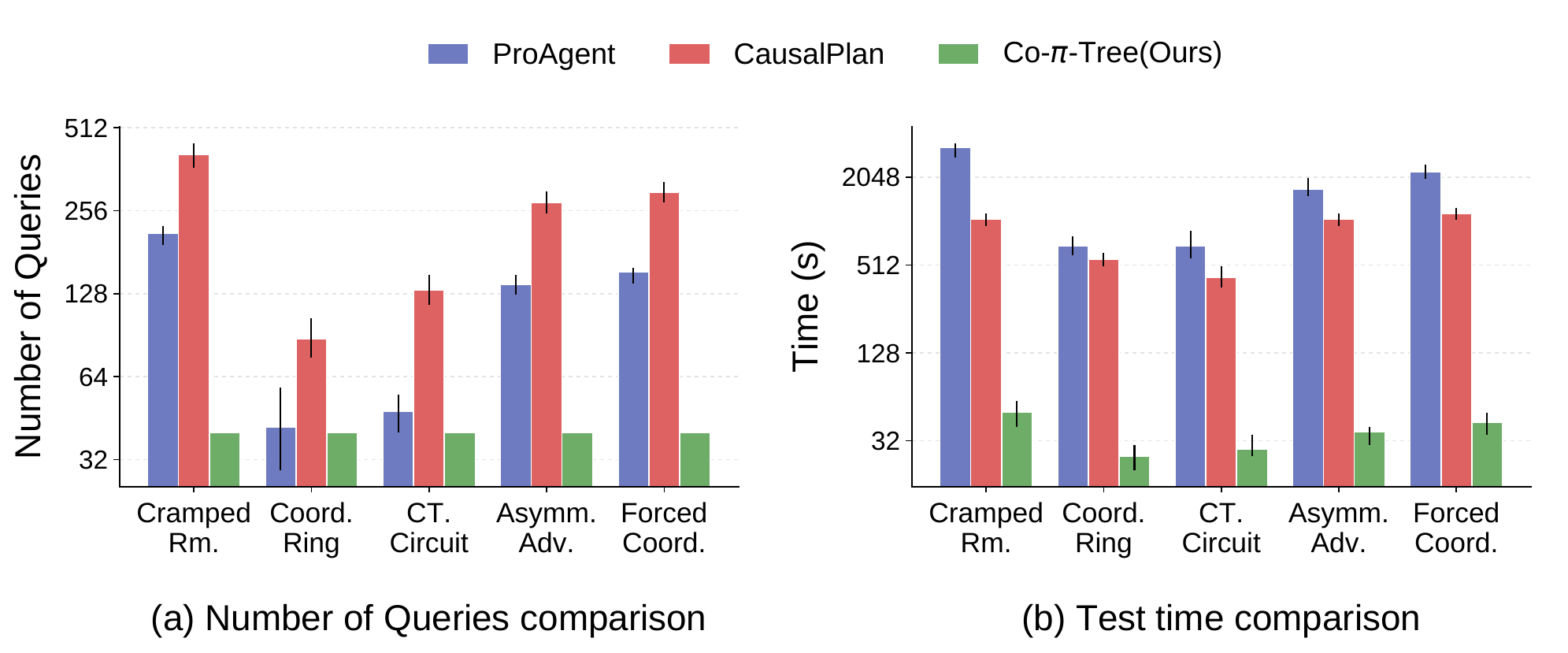}
\caption{Comparison among ProAgent, CausalPlan, and \method{} in terms of NQ and test-time latency.}
\label{fig:cost-comparison}
\end{figure}

Beyond reward, Figure~\ref{fig:cost-comparison} compares \method{} with
ProAgent and CausalPlan in terms of NQ and test-time latency. We repeat each evaluation five times and report averages. ProAgent and
CausalPlan rely on online test-time LLM reasoning, so each evaluation requires
repeated model queries and incurs response delay during interaction. For all
methods, NQ is counted for one complete algorithm run. Once the final policy
tree is produced, additional evaluations do not require extra LLM queries. As a
result, \method{} reduces NQ by 77.7\% and test-time latency by 97.1\% relative
to the mean of the two online LLM baselines.


\subsection{Collaborating with Human Partners}
To answer Q1 in the standard ZSC setting with real human partners, we further
evaluate whether the learned policy tree transfers to such partners. Humans interacted with the
evaluated agents through the human-AI web application of Overcooked-AI\footnote{\url{https://github.com/HumanCompatibleAI/overcooked_ai/tree/master/src/overcooked_demo}},
and we report average rewards. Detailed participant information, protocol details, and
method-selection rationale are
provided in Appendix~\ref{app:human}.

Figure~\ref{fig:human-box} shows the results for the
full \method{} and the two variants. The full method obtains
the highest mean reward on all five layouts, followed by \method{}-PI and then
\method{}-w/o P. Here, $p$ denotes the $p$-value of the Holm-corrected
two-sided Mann--Whitney U test. Smaller $p$ values and
more stars indicate stronger evidence that \method{} differs from the compared
method, while n.s. indicates no significant difference. Together with the higher
mean rewards, these markers support that the full method is stronger in most
human-AI collaboration comparisons.

These results suggest that partner-behavior prediction improves human-AI
coordination, and that explicitly conditioning action selection on the predicted
partner behavior further strengthens this effect. For example, when a soup is
ready, predicting that the human partner will deliver it allows the agent to
prepare onions for the next pot instead of duplicating the delivery task.

\subsection{Ablation Studies}

\noindent\textbf{Effect of Partner Prediction.}
Table~\ref{tab:ablation-partner} compares the full method with
variants.
\method{}-w/o P removes partner-behavior prediction entirely and is consistently
worse than \method{}-PI, indicating that partner
reasoning helps the agent predict partner behaviors and adjust its own actions
for coordination. At the same time, \method{}-PI often
performs best collaborating with other AI partners. This suggests that explicitly
conditioning action selection on partner-behavior prediction may be less
reliable when the partner's actions are hard to interpret or predict. 

\begin{table}[t]
\begin{nolinenumbers}
\centering
\small
\renewcommand{\arraystretch}{1.3}
\setlength{\tabcolsep}{1.5pt}
\begin{tabular}{lccc}
\hline
 & \method{} & \method{}-PI & \method{}-w/o P \\
\hline
Cramped Rm. & \textbf{182.3}\stdev{18} & \underline{176.7}\stdev{24} & 168.0\stdev{25} \\
 & \textbf{180.1}\stdev{16} & \underline{176.0}\stdev{19} & 165.7\stdev{22} \\
\hline
Coord. Ring & \underline{165.9}\stdev{28} & \textbf{168.0}\stdev{26} & 158.0\stdev{29} \\
 & \underline{162.1}\stdev{24} & \textbf{169.3}\stdev{23} & 152.0\stdev{25} \\
\hline
CT. Circuit & \textbf{117.8}\stdev{15} & \underline{110.7}\stdev{16} & 106.0\stdev{18} \\
 & \textbf{116.0}\stdev{17} & \underline{114.0}\stdev{18} & 105.2\stdev{21} \\
\hline
Asymm. Adv. & \underline{274.6}\stdev{31} & \textbf{282.7}\stdev{35} & 266.7\stdev{34} \\
 & \underline{239.1}\stdev{18} & \textbf{242.0}\stdev{15} & 234.7\stdev{22} \\
\hline
Forced Coord. & \underline{62.7}\stdev{31} & \textbf{66.0}\stdev{24} & 62.0\stdev{28} \\
 & \underline{35.6}\stdev{22} & \textbf{44.1}\stdev{26} & 32.8\stdev{23} \\
\hline
\end{tabular}
\caption{Ablation on partner prediction. Each entry reports mean team
reward $\pm$ std.}
\label{tab:ablation-partner}
\end{nolinenumbers}
\end{table}

\noindent\textbf{Effect of Iterative Refinement.}
Table~\ref{tab:ablation-refine} evaluates the effect of policy refinement.
We construct an ablation variant, \method{} w/o R, which removes policy
refinement and uses only the initial prompt construction. The full method
improves most layouts, indicating that episode
verbal feedback helps the planner repair weak branches and improve the final
policy tree.
Detailed reward growth curves covering all
five layouts and different partners are provided in
Appendix~\ref{app:iterative}.

\begin{table}[t]
\begin{nolinenumbers}
\centering
\small
\renewcommand{\arraystretch}{1.3}
\setlength{\tabcolsep}{5pt}
\begin{tabular}{lcc}
\hline
 & \method{} & \method{} w/o R \\
\hline
Cramped Rm. & \textbf{182.3}\stdev{18} & 163.1\stdev{16} \\
 & \textbf{180.1}\stdev{16} & 161.5\stdev{22} \\
\hline
Coord. Ring & \textbf{165.9}\stdev{28} & 153.8\stdev{29} \\
 & \textbf{162.1}\stdev{24} & 148.3\stdev{27} \\
\hline
CT. Circuit & \textbf{117.8}\stdev{15} & 104.0\stdev{18} \\
 & \textbf{116.0}\stdev{17} & 100.7\stdev{16} \\
\hline
Asymm. Adv. & \textbf{274.6}\stdev{31} & 260.3\stdev{25} \\
 & \textbf{239.1}\stdev{18} & 226.6\stdev{22} \\
\hline
Forced Coord. & 62.7\stdev{31} & \textbf{66.8}\stdev{29} \\
 & 35.6\stdev{22} & \textbf{35.8}\stdev{23} \\
\hline
\end{tabular}
\caption{Ablation on iterative refinement. Each entry reports mean team
reward $\pm$ std.}
\label{tab:ablation-refine}
\end{nolinenumbers}
\end{table}

\subsection{Additional Experiments}
We further conduct a cross-layout transfer experiment to evaluate scalability.
The results show that the learned policy tree retains strong transfer
performance without additional test-time LLM queries. Details are provided in
Appendix~\ref{app:layout-transfer}. We also provide a visualization analysis of
local policy tree refinement. The results show that the summarizer can identify
and refine a damaging branch, improving the accepted policy from the initial
policy. Details are provided in Appendix~\ref{app:interp}.

\section{Conclusion}
In this paper, we presented \method{}, a closed-loop policy learning algorithm
that distills LLM reasoning into interpretable
policy tree structures for human-AI collaboration. By learning a
two-component policy tree that predicts partner behavior and selects the agent's action, and by using grounded rollout feedback to revise problematic branches, \method{} moves LLM reasoning from
test-time control to policy learning while retaining partner-conditioned
coordination. Experiments in Overcooked-AI involving both unseen AI
partners and human partners show that the learned policies achieve strong ZSC performance, while requiring substantially fewer LLM
queries and lower test-time latency than online
LLM-based collaboration methods. Our results further suggest that reasoning
about a partner is broadly useful. Overall, \method{} provides a practical path toward efficient, auditable, and
generalizable collaborative agents.

\section*{Limitations}
This work has two limitations. (1) Although we evaluate \method{} with both AI partners and human partners in
Overcooked-AI, its effectiveness in physically embodied collaboration settings
remains to be further validated.
(2) \method{} currently relies on a manually defined action space and an
executor that grounds tree output actions into executable environment actions.
Future work may explore more automatic and adaptive action grounding methods,
reducing the need for specific executor design.



\bibliography{sample}
\clearpage
\appendix
\section{Additional Details}
\label{app:details}

\subsection{Layout Descriptions}
\label{app:layouts}

\begin{figure*}[t]
\centering
\includegraphics[width=0.8\textwidth]{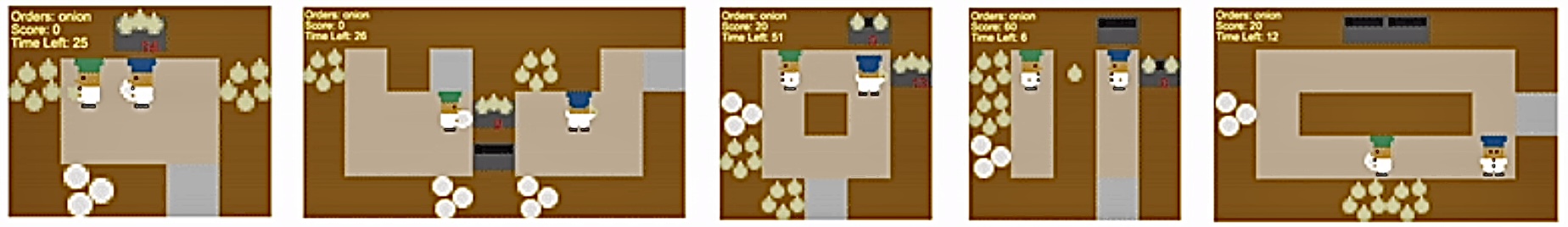}
\caption{Overcooked-AI layouts used in our evaluation.}
\label{fig:overcooked-layouts}
\end{figure*}

We evaluate \method{} on five standard Overcooked-AI layouts, which cover
different coordination bottlenecks.
Cramped Room is a compact kitchen with one pot and one serving
location. Because both players operate in a small shared area, good performance
requires avoiding blocking while keeping the single pot continuously in use.
Asymmetric Advantages places the two players in kitchens with
different access costs: one side is closer to onions and the other is closer to
serving-related resources. This layout rewards stable role specialization and
timely handoff between ingredient preparation and soup delivery.
Coordination Ring is a ring-shaped layout in which players must move
around narrow corridors and avoid obstructing each other. Efficient play usually
requires both pots to be used and for the players to maintain compatible
movement directions.
Forced Coordination separates the resources needed for cooking across
the two sides of the map. One side has access to ingredients and dishes, while
the other side handles pot interaction and serving, so successful completion
depends on passing objects through counters and performing complementary
subtasks. This makes Forced Coordination especially sensitive to player roles
and handoff conventions.
Counter Circuit is a larger ring-like layout with onions, dishes,
pots, and serving locations distributed across different regions. Its narrow
passages make movement conflicts frequent, and high reward often requires
placing intermediate objects on counters so that the other player can continue
the cooking pipeline. Figure~\ref{fig:overcooked-layouts} visualizes the five
layouts used in our experiments.

\subsection{Baseline Selection Details}
\label{app:baselines}

For the main AI-partner experiment, we include standard MARL and BC baselines
commonly used in Overcooked-AI ZSC evaluation, together with two LLM-based
baselines, ProAgent and CausalPlan. CausalPlan was under public review at the
time of our experiments. We evaluate the public May 2026 OpenReview/TMLR
version as a snapshot baseline, rather than treating either the earlier ICLR
submission or the May 2026 revision as a final archival version.

We also considered other recent LLM-based collaborators, including the
LLM-powered hierarchical language agent of \citet{Liu2023LLMPoweredHL}. That
method targets real-time human-AI coordination with natural-language commands:
its agent interprets human messages, maintains dialogue context, and maps
commands to macro actions. Its testbed is also a modified gym-cooking
environment with additional mechanics such as multiple ingredients, chopping,
order timeouts, fire, and a chat interface. Directly comparing it under our
Overcooked-AI ZSC protocol would therefore conflate policy quality with access
to an extra communication channel and different environment dynamics, so we do
not include it as a baseline. More generally, methods are excluded when they
lack public implementations or use incompatible task settings.

\subsection{Artifact Licenses}
\label{app:artifact-licenses}

We use public research artifacts under their stated licenses or terms of use.
The Overcooked-AI environment and Stable-Baselines components used in our
implementation are released under MIT licenses, and we retain their license
notices in the supplementary material. Public baseline implementations are used
for experimental comparison according to their accompanying licenses or public
release terms. LLM calls are made through the corresponding model provider
interface under its terms of use. The code and generated policy-tree artifacts
distributed with this paper include the applicable license and attribution
information.

\subsection{Data Privacy and Content}
\label{app:data-privacy}

Our experiments use Overcooked-AI simulator states, actions, rewards, and
execution traces. These records contain game-state information such as held
objects, pot states, selected actions, and execution outcomes, but do not contain
free-form participant messages or naturally occurring text. For the human study,
we assign anonymous participant identifiers and report only aggregate rewards
and coarse demographic statistics. We do not include names, contact information,
or other personally identifying information in the paper or supplementary
artifacts. Because the task domain is a constrained cooking game and the stored
traces use fixed symbolic fields rather than open-ended user text, the data do
not contain offensive content. Before release, we exclude any metadata that
could identify individual participants.

\subsection{AI-Partner Evaluation Protocol}
\label{app:ai-protocol}

For AI-partner evaluation, we consider both player role assignments in each
layout: the evaluated agent as Player 0 with the partner as Player 1, and the
swapped role setting. The unseen partner pool is
\{SP, PBT, FCP, MEP, COLE, BC\}. For each baseline in this six-partner pool,
we evaluate it with the other five held out partners and report the average
pairwise team reward. ProAgent and CausalPlan are evaluated against the same
six-partner pool and averaged in the same way.

For \method{} and its ablation variants, the held out test partners are never
used during policy construction, execution feedback, or refinement. Within
each layout, we select one partner as the source partner for policy
construction and refinement, and then evaluate the learned policy with the
remaining partners held out from that process. Results are averaged over all
source partner choices, so each number reflects zero-shot coordination with
unseen partners from the same reference pool. For each algorithm pair and role
assignment, we run five evaluation episodes and collect episode returns. Across
all tables, rewards are reported as mean$\pm$std. Bold indicates the best
result in each row and underlining indicates the second-best result.

\subsection{Forced Coordination Analysis}
\label{app:forced-coordination}

The results on Forced Coord. are more sensitive to player roles than those on
the other layouts. This is expected from the design of the layout: the two
players have access to different resources, and each soup requires successful
object passing through counters. As a result, a policy tree learned with one
source partner can fit that partner's handoff habit, such as when to leave an
onion or dish on a counter. When a held out partner follows a different habit,
the same branch may still choose legal actions but produce lower reward. We
therefore treat this phenomenon as a limitation of partner convention matching
in Forced Coord., rather than as a failure of the executor or the policy tree
format.

\subsection{Policy-Learning Details}
\label{app:algorithm}

Algorithm~\ref{alg:codec-tree} in the main methodology summarizes the
\method{} policy-learning loop. This section provides the implementation and
parameter details omitted from the algorithm block.

\paragraph{LLM settings and temperature schedule.}
Unless otherwise stated, \method{} runs for 10 refinement iterations. All LLM
modules use the same GPT-4o backbone. The planner and coder use the
stuck-aware temperature schedule shown below. Let $u_k$ denote the stuck
counter, which increases when a newly generated policy fails to improve over
the best accepted policy. The sampling temperature for planner and coder calls
is
\begin{equation}
    \vartheta_k = \min\!\left(\vartheta_{\max},
    \vartheta_{\mathrm{base}}\left(1 + \log(1 + u_k)\right)\right),
\end{equation}
where $\vartheta_{\mathrm{base}}=0.4$ and $\vartheta_{\max}=1.2$. This schedule keeps
early iterations relatively stable while gradually encouraging broader
exploration when refinement becomes stuck. The summarizer uses the same
backbone to convert the accepted rollout trace into a compact policy summary
and a localized policy tree reflection. We set the maximum generation length to
2048 tokens for the policy tree generation, code-generation, and summarization
calls.

\subsection{Computational Details and Package Settings}
\label{app:computational-details}

\method{} does not train or fine-tune neural networks. Its policy-learning
process uses GPT-4o through an API to generate and refine an executable policy
tree, and the final policy is ordinary Python code executed without LLM calls
at test time. The exact number of GPT-4o parameters and the provider-side
inference infrastructure are not publicly disclosed. We therefore report the
controllable computation in terms of LLM query count (NQ) and test-time latency
in Section~\ref{sec:experiments} and Figure~\ref{fig:cost-comparison}. Local
computation consists of Overcooked-AI simulator rollouts, deterministic
executor calls, and policy-tree execution; no local GPU training is used for
\method{}, so the local training GPU budget for our method is 0 GPU-hours.

We use the public Overcooked-AI implementation for environment dynamics,
layouts, rewards, and the web-based human-AI interface. Episodes last 400
environment steps, and we evaluate the five layouts, partner pool, and
role assignment protocol described in Section~\ref{sec:experiments} and
Appendix~\ref{app:ai-protocol}. Baseline policies are evaluated with their
public implementations, checkpoints, and default environment interfaces when
available. For \method{}, all variants use the same symbolic state schema,
action vocabulary, deterministic executor, LLM backbone, temperature schedule,
number of refinement iterations, and maximum generation length described in
Appendix~\ref{app:algorithm} and Appendix~\ref{app:grounding-interfaces}; we
did not perform a separate hyperparameter search.

\subsection{Environment Grounding}
\label{app:environment-grounding}

This section details the environment grounding interface used to connect the
generated policy tree with Overcooked-AI execution. The raw simulator state
contains player positions and orientations, held objects, object locations,
pot contents, cooking timers, and layout-specific counter information. Before
calling the generated policy, the executor extracts the task-relevant symbolic
state and stores it in a fixed dictionary:
\begin{policybox}{Symbolic State Dictionary}
\begin{Verbatim}[fontsize=\tiny,breaklines=true,breakanywhere=true,breaksymbolleft={},breaksymbolright={}]
state_dict = {
    'hold': [None, None],  # [player0_hold, player1_hold], each in {'empty','onion','dish','soup'}
    'pot': [],             # list of dicts: {'count': int, 'state': 'idle'|'cooking'|'ready', 'timers': int|None}
    'any_soup_ready': None,
    'any_pot_not_full': None,
    'teammate_last_completed_skill': None,
    'teammate_last_inferred_skill': None,
    'self_last_skill': None,
    # optional for Forced Coordination:
    'num_empty_counters': None,
    'num_onion_counters': None,
    'num_dish_counters': None,
}
\end{Verbatim}
\end{policybox}

The generated Python policy receives this dictionary and returns two symbolic
decisions: the predicted teammate behavior and the selected action for the
controlled agent. Both are represented with the same action vocabulary. The
selected action is then passed to a deterministic executor. For
object-manipulation actions such as
\texttt{pickup\_onion}, \texttt{put\_onion\_in\_pot},
\texttt{pickup\_dish}, \texttt{fill\_dish\_with\_soup},
\texttt{deliver\_soup}, and \texttt{place\_obj\_on\_counter}, the executor
selects feasible interaction targets, computes the next executable environment
action, and records whether the action could be executed in the current state.

The same grounding layer also constructs the rollout trace $\tau_k$ used by
the summarizer. Each trace entry keeps the symbolic dictionary together with a
compact language description of the scene, the policy-selected action, the
execution outcome, and, in AI-partner rollouts, the partner's realized
action. For example, a trace entry may describe that the controlled
agent holds an onion, the teammate holds a dish, one pot is cooking with three
onions, another pot still needs onions, and the selected action was executable.
This representation gives the summarizer scene-level evidence for local branch
revision while avoiding the cost of storing full tensor trajectories in the
LLM context.

\subsection{Grounding Interfaces and Executor}
\label{app:grounding-interfaces}

Baselines are evaluated with their own environment grounding interfaces, as
provided by their implementations or required by their action spaces. For
\method{}, the executor is a deterministic grounding layer: given a tree output
action selected by the policy tree, it chooses a reachable target and outputs
the next executable environment action. It does not choose the tree output action,
query the LLM, use reward feedback, or adapt to partner identity. Our executor
extensions mainly support the policy-tree state schema, stable target
selection, feasibility checks, and execution-trace logging. All \method{}
variants use the same executor, so ablation results isolate partner prediction
and iterative refinement rather than executor design.

\subsection{Human Study Protocol}
\label{app:human}

Following prior Overcooked human-agent evaluation protocols, we recruited
20 volunteers from a local university, including 9 female and 11 male
participants, with ages ranging from 18 to 30, and assigned them to layout
conditions. Nearly all participants were unfamiliar with Overcooked before the
study. We therefore provided comprehensive task instructions and allowed each
participant to complete at least five practice rounds before evaluation.
Before the study, participants were told that the purpose was to evaluate
human-AI collaboration in a cooperative cooking game, that their gameplay
records would be used only for research analysis, and that only anonymized and
aggregate results would be reported. The instructions explained the game goal,
basic controls, evaluation procedure, expected duration, compensation, and that
there were no known risks beyond ordinary computer-game interaction.
Participation was voluntary, and participants provided consent before starting
the practice rounds.
The order of evaluated agents was randomized within each layout to reduce order
effects.
Participants then interacted with the evaluated agents through the web-based
human-AI interface of \citet{Carroll2019OnTU}. For each evaluated method, each
volunteer completed two evaluation episodes under the two player-role
assignments, and we recorded the average reward over the two episodes. In the
main paper, the box plots aggregate these per-volunteer average rewards, so
the two role assignments are merged rather than reported separately.
Participants were recruited through local university channels and received a
fixed payment for their time, independent of game score. The payment amount was
set according to local campus participation norms and the short duration of the
study.

We do not include BC in the human-partner study because BC is used as a proxy
model of human behavior rather than as a real-time collaborative policy for
direct human interaction. We also exclude ProAgent because its action-by-action
online LLM querying introduces prohibitive response latency in the real-time
web interface used for human evaluation.

\subsection{Iterative Reward Growth}
\label{app:iterative}

\begin{figure*}[t]
\centering
\includegraphics[width=\textwidth]{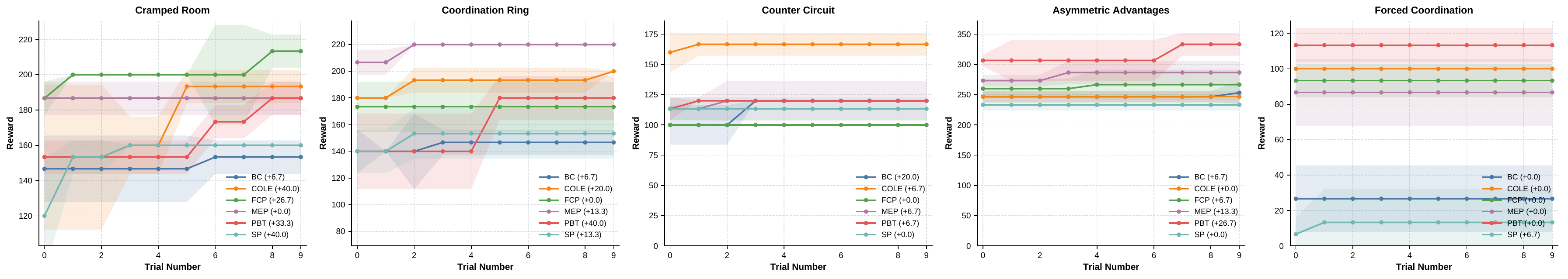}
\caption{Reward growth over iterative refinement across five Overcooked-AI
layouts. Each panel shows one representative training run with a different
source partner. We plot the accepted reward trajectory across 10 refinement
iterations; shaded bands indicate one standard deviation over evaluation
episodes.}
\label{fig:iterative-growth}
\end{figure*}

Figure~\ref{fig:iterative-growth} visualizes reward growth during iterative
refinement across all five layouts. Each panel corresponds to one layout-source
partner setting and reports the accepted reward trajectory over 10 refinement
iterations. The curves show that accepted policy performance often improves
over the early iterations and then stabilizes, which is consistent with the
accept/revert mechanism in Algorithm~\ref{alg:codec-tree}. The trend is most
visible in layouts where local branch repairs can immediately improve role
division or object handoff. In more partner-sensitive layouts, such as
Forced Coord., the trajectory can be flatter because final reward also
depends heavily on whether the partner completes the complementary subtasks
required by the layout.

\subsection{Cross-Layout Transfer}
\label{app:layout-transfer}

We further evaluate whether a policy learned on one layout can be reused on a
different layout without additional LLM calls. For \method{}, a policy tree is
first learned on a \emph{source layout} with BC as the source partner and is
then directly deployed on a different \emph{target layout}. During transfer,
the policy code is reused as is: we do not query the LLM, regenerate the tree,
refine the policy, or tune any layout-specific parameters. The only cost during
target-layout evaluation is ordinary policy-tree execution. We compare this
zero-extra-LLM transfer setting with ProAgent, which performs online LLM
reasoning on each target layout, and with SP as a representative MARL baseline
that also has zero LLM cost at test-time.

Because different Overcooked-AI layouts have different reward scales, we report
normalized transfer retention rather than raw reward. Let $R_{s\rightarrow t}$
denote the mean reward obtained by a method trained or instantiated on source
layout $s$ and evaluated on target layout $t$. For a source layout $s$ and
target layout $t$, the cell value is computed as
\[
    \mathrm{Retention}(s,t)
    =
    \frac{R_{s\rightarrow t}}{R_{t\rightarrow t}}
    \times 100\%.
\]
Thus, the diagonal is 100\% by definition, and off-diagonal entries measure
how much of the target-layout in-domain performance is retained under direct
cross-layout test-time.

\begin{figure*}[t]
\centering
\includegraphics[width=\textwidth]{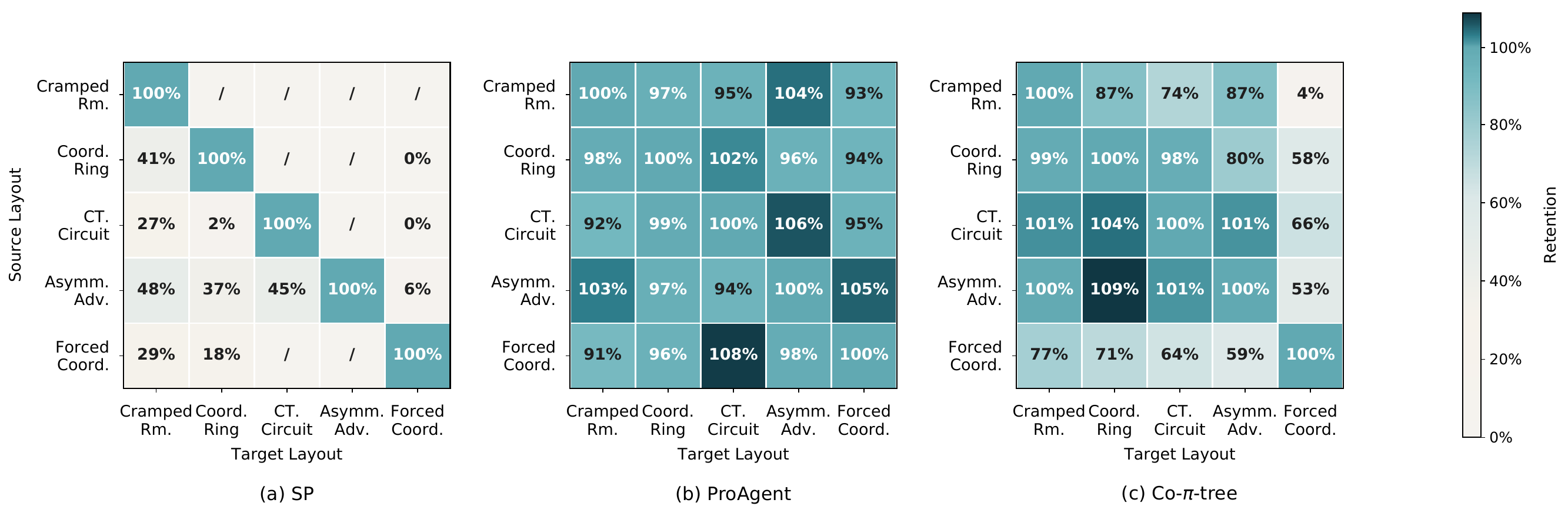}
\caption{Cross-layout transfer performance for SP, ProAgent, and \method{}.
Rows denote the source layout used to learn or instantiate the policy, and
columns denote the target layout used for evaluation. Each cell reports
normalized transfer retention, computed as the target-layout reward of a
source-layout policy divided by the in-domain reward of the corresponding
target-layout policy. Percentages make results comparable across layouts with
different reward scales. For SP, ``/'' denotes source-target pairs that cannot
be directly executed because of incompatible layout-specific tensor
observations; compatible pairs are evaluated with the zero-padding procedure
described in the text.}
\label{fig:layout-transfer}
\end{figure*}

For SP, only a subset of source-target layout pairs can be executed directly in
our current Overcooked implementation. SP consumes layout-specific tensor
observations whose spatial shapes depend on the map geometry, so a source
policy cannot generally be applied to a target layout whose observation tensor
is incompatible with the source input shape. We therefore evaluate the
compatible subset, using zero-padding when needed to embed smaller target
observations into the source policy's expected tensor shape. Cells that still
cannot be executed are marked with ``/'' in Figure~\ref{fig:layout-transfer}
and are shown with the lowest heatmap color only for visualization; cells
annotated as 0\% are executable pairs whose measured retention is near zero.

Figure~\ref{fig:layout-transfer} shows three different transfer behaviors.
SP transfers poorly even on the compatible subset, confirming that the neural
policy is strongly coupled to the training layout. ProAgent achieves the best
cross-layout retention because it reasons online and can adapt its plan to the
target layout, but this requires high test-time LLM consumption on every run.
In contrast, \method{} retains strong transfer performance with no additional
LLM calls at transfer time. Several off-diagonal cells are close to or above
100\%, indicating that the learned policy tree is not merely memorizing a
single layout but captures transferable subtask structure such as ingredient
preparation, dish handling, delivery, and partner-complementary role selection.
Transfer remains weaker for Forced Coord., where success depends heavily on
layout-specific object passing and on whether the partner completes the
required complementary subtask.

\subsection{Interpretability and Local Policy Tree Refinement}
\label{app:interp}

Because \method{} stores its policy as explicit branches, both the initial
policy and the refined policy can be inspected directly. Below we show an
Overcooked-AI example from Cramped Room with BC as the source partner. The
initial policy obtains an average score of 160.0. In this Overcooked-AI case,
the summarizer identifies a local dish-handling branch that can cause premature
dish pickup when pot preparation is still useful. After refinement, the best
accepted policy obtains an average score of 186.7.
In the displayed trees, \textcolor{red!75!black}{red} marks the original
branches targeted by accepted local reflections, and
\textcolor{green!45!black}{green} marks the corresponding repaired branches.

\paragraph{Initial Policy Tree.}
The full initial policy tree is shown below.

\begin{policybox}{Initial Policy Tree}
\begin{PolicyVerb}
### FunctionDescription:
  Name: PredictTeammateThenPlan
  Inputs:
    - current_scene:
        - holdings: {self: empty/onion/dish/soup, teammate: empty/onion/dish/soup}
        - pots: for each <Pot>: onion_count: {0,1,2,3}, state: {idle,cooking,ready}, timers
        - derived flags: any_soup_ready, any_pot_not_full
    - teammate_last_completed_skill: one of the 6 actions or None
    - teammate_last_inferred_skill: one of the 6 actions or None
    - self_last_skill: one of the 6 actions or None
  Outputs:
    - inferred_teammate_current_skill: one of the 6 actions
    - self_action_now: one of the 6 actions

  **Partner-Inference Policy Tree**
  If `teammate_last_completed_skill` is `pickup_onion` and `teammate` is holding an onion:
      - inferred_teammate_current_skill = `put_onion_in_pot`
  If `teammate_last_completed_skill` is `put_onion_in_pot` and `teammate` is holding nothing:
      - inferred_teammate_current_skill = `pickup_onion`
  If `teammate_last_completed_skill` is `pickup_dish` and `teammate` is holding a dish:
      - inferred_teammate_current_skill = `fill_dish_with_soup`
  If `teammate_last_completed_skill` is `fill_dish_with_soup` and `teammate` is holding soup:
      - inferred_teammate_current_skill = `deliver_soup`
  If `teammate_last_completed_skill` is `deliver_soup` and `teammate` is holding nothing:
      - inferred_teammate_current_skill = `pickup_dish`
  If `teammate` is holding an onion and no recent action suggests otherwise:
      - inferred_teammate_current_skill = `put_onion_in_pot`
  If `teammate` is holding a dish and no recent action suggests otherwise:
      - inferred_teammate_current_skill = `fill_dish_with_soup`
  If `teammate` is holding soup and no recent action suggests otherwise:
      - inferred_teammate_current_skill = `deliver_soup`
\end{PolicyVerb}
\begin{PolicyVerbRed}
  If `teammate` is holding nothing and no recent action suggests otherwise:
      - inferred_teammate_current_skill = `pickup_onion`
\end{PolicyVerbRed}
\begin{PolicyVerb}

Self-Action Policy Tree

If `self_hold` is `soup`:
    - self_action_now = `deliver_soup`

\end{PolicyVerb}
\begin{PolicyVerbRed}
If `self_hold` is `dish`:
    - If `current_scene['any_soup_ready']`:
        - self_action_now = `fill_dish_with_soup`
    - Else:
        - self_action_now = `place_obj_on_counter`
\end{PolicyVerbRed}
\begin{PolicyVerb}

If `self_hold` is `onion`:
    - If there exists a `pot` that is not full:
        - self_action_now = `put_onion_in_pot`
    - Else:
        - self_action_now = `place_obj_on_counter`

If `self_hold` is `empty`:
    - If `current_scene['any_pot_not_full']`:
        - self_action_now = `pickup_onion`
    - Else:
        - If `current_scene['any_soup_ready']`:
            - self_action_now = `pickup_dish`
        - Else:
            - self_action_now = `pickup_onion`

If `current_scene['any_pot_not_full']` is `True` and `self_hold` is `empty`:
    - self_action_now = `pickup_onion`

\end{PolicyVerb}
\begin{PolicyVerbRed}
If `current_scene['any_soup_ready']` is `True` and `self_hold` is `empty`:
    - self_action_now = `pickup_dish`
\end{PolicyVerbRed}
\begin{PolicyVerb}

If `current_scene['any_pot_not_full']` is `False` and `current_scene['any_soup_ready']` is `False`:
    - If `self_hold` is `empty`:
        - self_action_now = `pickup_onion`

If `self_hold` is `empty` and the teammate's inferred action is `pickup_onion`:
    - self_action_now = `pickup_onion`

If `self_hold` is `empty` and the teammate's inferred action is `put_onion_in_pot`:
    - self_action_now = `pickup_onion`

If `self_hold` is `empty` and the teammate's inferred action is `pickup_dish`:
    - If `current_scene['any_soup_ready']`:
        - self_action_now = `pickup_onion`
    - Else:
        - self_action_now = `pickup_dish`

If `self_hold` is `empty` and the teammate's inferred action is `fill_dish_with_soup`:
    - If `current_scene['any_pot_not_full']`:
        - self_action_now = `pickup_onion`
    - Else:
        - self_action_now = `pickup_dish`

If `self_hold` is `empty` and the teammate's inferred action is `deliver_soup`:
    - If `current_scene['any_pot_not_full']`:
        - self_action_now = `pickup_onion`
    - Else:
        - self_action_now = `pickup_dish`

If `self_hold` is `empty` and the teammate's inferred action is `place_obj_on_counter`:
    - self_action_now = `pickup_onion`
\end{PolicyVerb}
\end{policybox}

\paragraph{Policy Tree reflection.}
The refinement run generates multiple candidate reflections. We report only the
three local reflections whose \texttt{After\_Behavior} is directly absorbed by
the later accepted policies and is visible in the final refined tree. We omit
suggestions from unsuccessful resampling or from later unused proposals.

\begin{policybox}{Accepted Local Reflections}
\begin{Verbatim}[fontsize=\tiny,breaklines=true,breakanywhere=true,breaksymbolleft={},breaksymbolright={}]
Reflection 1
Target_Branch: Self Action Selection Tree - Condition 4 and 6
Modification_Scope: LOCAL_SINGLE_BRANCH
Before_Behavior: If `self_hold` is `empty` and `current_scene['any_soup_ready']` is `True`, the player always selects `pickup_dish` as the next action, even when `current_scene['any_pot_not_full']` is also `True`.
After_Behavior:
  - If `self_hold` is `empty` and `current_scene['any_soup_ready']` is `True`, the player selects `pickup_dish` ONLY if `current_scene['any_pot_not_full']` is `False`. Otherwise, prioritize `pickup_onion` to prepare the next cooking cycle.
Unchanged_Assumptions:
  - The logic for handling soup delivery (`self_hold` is `soup`) remains unchanged.
  - The logic for handling onion placement (`self_hold` is `onion`) remains unchanged.
  - The teammate-behavior prediction process and its integration into self-action selection remain unchanged.
  - The overall prioritization of tasks based on game state flags (`any_soup_ready`, `any_pot_not_full`) remains intact.
Expected_Effect: By refining the condition to avoid premature dish pickup, the player will focus on preparing onions when pots are not full, reducing idle time and ensuring a smoother cooking cycle. This change is expected to eliminate redundant dish handling and improve overall efficiency, potentially increasing the game score.

Reflection 2
Target_Branch: Self Action Selection Tree, Rule 2 (If self_hold is dish)
Modification_Scope: LOCAL_SINGLE_BRANCH
Before_Behavior: If self_hold is dish, the logic checks whether any soup is ready. If soup is ready, the player fills the dish with soup. If no soup is ready, the player either places the dish on the counter or picks up an onion depending on the state of pots.
After_Behavior:
  - If self_hold is dish:
  - 1. Check if any soup is ready.
    - If soup is ready, self_action_now = fill_dish_with_soup.
  - 2. If no soup is ready, check if any pot is cooking or idle.
    - If any pot is cooking, self_action_now = place_obj_on_counter.
    - If all pots are idle, self_action_now = pickup_onion.
Unchanged_Assumptions:
  - The logic for handling actions when self_hold is empty remains unchanged.
  - The predicted teammate behavior logic remains unchanged.
  - The priority of filling dishes with soup when soup is ready remains unchanged.
  - The rules for handling onions and pots remain unchanged.
Expected_Effect: This modification prevents premature dish pickup and holding while pots are still cooking, allowing the player to engage in more productive tasks like onion collection or pot preparation. It eliminates unnecessary holding time and improves overall efficiency.

Reflection 3
Target_Branch: Partner Inference Tree, Rule 9
Modification_Scope: LOCAL_SINGLE_BRANCH
Before_Behavior: If `teammate` is holding nothing and no recent action suggests otherwise, inferred_teammate_current_skill = `pickup_onion`.
After_Behavior:
  - If `teammate` is holding nothing and no recent action suggests otherwise:
    - If `current_scene['any_soup_ready']` is true and `current_scene['any_pot_not_full']` is false:
        inferred_teammate_current_skill = `pickup_dish`
    - Else:
        inferred_teammate_current_skill = `pickup_onion`
Unchanged_Assumptions:
  - The predicted teammate behavior is based on their last completed action, current holding state, and the current scene context.
  - The predicted teammate behavior is used to guide self-action prioritization.
  - The predicted teammate behavior cannot override explicit recent action observations (e.g., `teammate_last_completed_skill`).
  - The overall structure and order of other rules in the Partner Inference Tree remain unchanged.
Expected_Effect: This modification will reduce the misprediction of the teammate's behavior during the soup cooking phase, ensuring that the teammate is correctly predicted to prioritize picking up a dish for soup delivery when all pots are full and soup is ready. This will minimize redundant actions (e.g., picking up onions unnecessarily) and improve the overall efficiency of the team, leading to higher scores.
\end{Verbatim}
\end{policybox}

\paragraph{Example analysis.}
In this Overcooked-AI example, the initial tree has an over-eager
dish-handling pattern around the soup-ready state. When the agent is empty-handed and
\texttt{any\_soup\_ready} is true, it immediately chooses
\texttt{pickup\_dish}, even when \texttt{any\_pot\_not\_full} is also true and
onion preparation is still useful. The first reflection names this exact
empty-hand branch and changes the rule so that dish pickup is selected only
when no pot still needs ingredients; otherwise the agent continues with
\texttt{pickup\_onion}. The second reflection fixes the related case in which
the agent is already holding a dish but no soup is ready: instead of using a
single fallback action, the refined tree checks the pot state, puts the dish
down while a pot is cooking, and returns to onion preparation when all pots are
idle. The third reflection adjusts the partner-inference branch for an
empty-handed teammate. The initial tree always predicts
\texttt{pickup\_onion}; the refined tree predicts \texttt{pickup\_dish} when
soup is ready and no pot needs more onions. Together, these changes turn a
coarse dish-first heuristic into state-conditioned coordination between dish
delivery and the next onion cycle, increasing the average score from 160.0 to
186.7.

\paragraph{Refined Policy Tree.}
The full refined policy tree is shown below.

\begin{policybox}{Refined Policy Tree}
\begin{PolicyVerb}
### FunctionDescription:
  Name: PredictTeammateThenPlan
  Inputs:
    - current_scene:
        - holdings: {self: empty/onion/dish/soup, teammate: empty/onion/dish/soup}
        - pots: for each <Pot>: onion_count: {0,1,2,3}, state: {idle,cooking,ready}, timers
        - derived flags: any_soup_ready, any_pot_not_full
    - teammate_last_completed_skill: one of the 6 actions or None
    - teammate_last_inferred_skill: one of the 6 actions or None
    - self_last_skill: one of the 6 actions or None
  Outputs:
    - inferred_teammate_current_skill: one of the 6 actions
    - self_action_now: one of the 6 actions

  **Partner-Inference Policy Tree**
  If `teammate_last_completed_skill` is `pickup_onion` and `teammate` is holding an onion:
      - inferred_teammate_current_skill = `put_onion_in_pot`
  If `teammate_last_completed_skill` is `put_onion_in_pot` and `teammate` is holding nothing:
      - inferred_teammate_current_skill = `pickup_onion`
  If `teammate_last_completed_skill` is `pickup_dish` and `teammate` is holding a dish:
      - inferred_teammate_current_skill = `fill_dish_with_soup`
  If `teammate_last_completed_skill` is `fill_dish_with_soup` and `teammate` is holding soup:
      - inferred_teammate_current_skill = `deliver_soup`
  If `teammate_last_completed_skill` is `deliver_soup` and `teammate` is holding nothing:
      - inferred_teammate_current_skill = `pickup_dish`
  If `teammate` is holding an onion and no recent action suggests otherwise:
      - inferred_teammate_current_skill = `put_onion_in_pot`
  If `teammate` is holding a dish and no recent action suggests otherwise:
      - inferred_teammate_current_skill = `fill_dish_with_soup`
  If `teammate` is holding soup and no recent action suggests otherwise:
      - inferred_teammate_current_skill = `deliver_soup`
\end{PolicyVerb}
\begin{PolicyVerbGreen}
  If `teammate` is holding nothing and no recent action suggests otherwise:
      - If `current_scene['any_soup_ready']` is true and `current_scene['any_pot_not_full']` is false:
          - inferred_teammate_current_skill = `pickup_dish`
      - Else:
          - inferred_teammate_current_skill = `pickup_onion`
\end{PolicyVerbGreen}
\begin{PolicyVerb}

### Self-Action Policy Tree

If `self_hold` is `soup`:
    - self_action_now = `deliver_soup`

\end{PolicyVerb}
\begin{PolicyVerbGreen}
If `self_hold` is `dish`:
    - If `current_scene['any_soup_ready']`:
        - self_action_now = `fill_dish_with_soup`
    - Else:
        - If there exists a `pot` that is cooking:
            - self_action_now = `place_obj_on_counter`
        - If all pots are idle:
            - self_action_now = `pickup_onion`
\end{PolicyVerbGreen}
\begin{PolicyVerb}

If `self_hold` is `onion`:
    - If there exists a `pot` that is not full:
        - self_action_now = `put_onion_in_pot`
    - Else:
        - self_action_now = `place_obj_on_counter`

If `self_hold` is `empty`:
    - If `current_scene['any_pot_not_full']`:
        - self_action_now = `pickup_onion`
    - Else:
        - If `current_scene['any_soup_ready']`:
            - If `teammate_last_inferred_skill` is `pickup_dish` or `fill_dish_with_soup`:
                - self_action_now = `pickup_onion`
            - Else:
                - self_action_now = `pickup_dish`
        - Else:
            - self_action_now = `pickup_onion`

If `self_hold` is `empty` and `current_scene['any_pot_not_full']` is `True`:
    - self_action_now = `pickup_onion`

\end{PolicyVerb}
\begin{PolicyVerbGreen}
If `self_hold` is `empty` and `current_scene['any_soup_ready']` is `True`:
    - If `current_scene['any_pot_not_full']` is `False`:
        - self_action_now = `pickup_dish`
    - Else:
        - self_action_now = `pickup_onion`
\end{PolicyVerbGreen}
\begin{PolicyVerb}

If `current_scene['any_pot_not_full']` is `False` and `current_scene['any_soup_ready']` is `False`:
    - If `self_hold` is `empty`:
        - self_action_now = `pickup_onion`

If `self_hold` is `empty` and the teammate's inferred action is `pickup_onion`:
    - self_action_now = `pickup_onion`

If `self_hold` is `empty` and the teammate's inferred action is `put_onion_in_pot`:
    - self_action_now = `pickup_onion`

If `self_hold` is `empty` and the teammate's inferred action is `pickup_dish`:
    - If `current_scene['any_soup_ready']`:
        - If `current_scene['any_pot_not_full']`:
            - self_action_now = `pickup_onion`
        - Else:
            - self_action_now = `pickup_dish`
    - Else:
        - self_action_now = `pickup_onion`

If `self_hold` is `empty` and the teammate's inferred action is `fill_dish_with_soup`:
    - If `current_scene['any_pot_not_full']`:
        - self_action_now = `pickup_onion`
    - Else:
        - self_action_now = `pickup_dish`

If `self_hold` is `empty` and the teammate's inferred action is `deliver_soup`:
    - If `current_scene['any_pot_not_full']`:
        - self_action_now = `pickup_onion`
    - Else:
        - self_action_now = `pickup_dish`

If `self_hold` is `empty` and the teammate's inferred action is `place_obj_on_counter`:
    - self_action_now = `pickup_onion`
\end{PolicyVerb}
\end{policybox}

This example illustrates two forms of interpretability. First, the learned
policy exposes the reason for an action through a small number of symbolic
conditions. Second, refinement is local: the summarizer names the problematic
branch, describes the before/after behavior, and leaves unrelated branches
unchanged.

\subsection{Prompt Templates}
\label{app:prompts}

This section gives the prompt templates used by the three LLM modules in
\method{}. The planning prompt contains the task objective, task rules, the
legal action library, the symbolic input schema, and a small set of
demonstration scenes. Each demonstration specifies the structured scene
description, the predicted partner behavior, the selected self action, and a
short justification. The code prompt fixes the Python function signature, the input
dictionary fields, and the required return tuple. Across source partners, we
keep the prompt format and demonstration structure fixed, while task-specific
information enters through the structured scene description and task
description.

In our Overcooked-AI instantiation, layout-role prompts instantiate the
placeholders for the controlled player, teammate player, executable action
sets, and layout-specific symbolic fields. In most layouts, both players can
execute all six Overcooked-AI actions. In Forced Coordination, the prompt
additionally restricts the controlled player's action set for the separated side
and includes counter features such as \texttt{num\_empty\_counters},
\texttt{num\_onion\_counters}, and \texttt{num\_dish\_counters}.

Following Section~3.2, the planner output is organized into two components:
the \texttt{Partner Inference Tree} implements the partner-behavior prediction
tree $T^{\mathrm{pred}}$, and the \texttt{Self Action Selection Tree}
corresponds to $T^{\mathrm{act}}$. The template below shows the combined
\texttt{FunctionDescription} used to store the final policy tree
$T=(T^{\mathrm{pred}},T^{\mathrm{act}})$; in a split planner implementation,
the same schema is produced by generating the partner-behavior prediction
component and then continuing with the self-action component.

\paragraph{Planner prompt.}
The planner receives task knowledge, legal actions, the symbolic input schema,
demonstration scenes, and, after the first iteration, memory and reflection
from the previous accepted policy. The concrete template below is the
Overcooked-AI instantiation of this general prompt structure.

\begin{policybox}{Planner Prompt Template}
\begin{Verbatim}[fontsize=\tiny,breaklines=true,breakanywhere=true,breaksymbolleft={},breaksymbolright={}]
[Task and rule knowledge]
Instructions:
- The Overcooked_AI game requires two players to work together as a team with the goal of achieving the highest possible score.
- To get points, the team must make soup according to the recipe, fill the soup in a dish, and immediately deliver the soup. Once a delivery is made, the team gets 20 points. The soup and dish then disappear.
- Recipe: three onions in the <Pot>.
- To make a soup, the team must pick up three onions one by one and put them in a <Pot>. The <Pot> automatically starts cooking when it contains three onions, and cooking takes 20 timesteps.
- Each player can hold at most one item. To put down the item a player is holding and empty the hand, use place_obj_on_counter.
- <Pot> can ONLY hold three ingredients.
- After cooking starts, before the soup is finished:
  - If no soup is ready, do NOT pick up a dish.
  - If there is a <Pot> not full in the environment, prepare for another cooking cycle.

[Legal actions]
In this game, each player can ONLY perform the following allowed actions. Do not use any other actions.
- pickup_onion
  - I need to have nothing in hand.
- put_onion_in_pot
  - I need to have an onion in hand.
- pickup_dish
  - Need to have a soup ready in <Pot>.
  - I need to have nothing in hand.
  - If there is no ready soup in the current scene, I should not pickup_dish.
- fill_dish_with_soup
  - I must do fill_dish_with_soup when soup is ready in <Pot>.
  - I need to have a dish in hand.
  - Then I must deliver_soup.
- deliver_soup
  - I must do deliver_soup after fill_dish_with_soup, when I hold soup.
  - I need to have soup in hand.
  - The dish and soup disappear after delivery.
- place_obj_on_counter
  - I need to have something in hand.
  - Do not use place_obj_on_counter when I hold soup.

Each player may only choose a subset of these actions:
- <SELF_PLAYER> can execute: <SELF_SKILL_SET>
- <TEAMMATE_PLAYER> can execute: <TEAMMATE_SKILL_SET>
<LAYOUT_SPECIFIC_RULES_AND_FIELDS>

[Role and output task]
Suppose you are an assistant proficient in the Overcooked_AI game. Your goal is to control <SELF_PLAYER> and cooperate with <TEAMMATE_PLAYER>, who is controlled by a certain strategy, to get a high score.
- <SELF_PLAYER> and <TEAMMATE_PLAYER> cannot communicate.
- You cannot use move actions and do not use location information in the observation.
- You must do deliver_soup when you hold soup.
- If there is no ready soup in the current scene, you should not pickup_dish.
- You need to define ONE function called every timestep. It receives the current scene and returns:
  1. The inferred current action of <TEAMMATE_PLAYER>, and
  2. Exactly one legal action for <SELF_PLAYER> now.

Required output format:
### FunctionDescription:
  Name: PredictTeammateThenPlan
  Inputs:
    - current_scene:
        - holdings: {self: empty/onion/dish/soup, teammate: empty/onion/dish/soup}
        - pots: for each <Pot>: onion_count: {0,1,2,3}, state: {idle,cooking,ready}, timers
        - derived flags: any_soup_ready, any_pot_not_full
        - optional layout fields: num_empty_counters, num_onion_counters, num_dish_counters
    - teammate_last_completed_skill: one of the legal actions or None
    - teammate_last_inferred_skill: one of the legal actions or None
    - self_last_skill: one of the legal actions or None
  Outputs:
    - inferred_teammate_current_skill: one legal teammate action
    - self_action_now: one legal self action

  **Partner Inference Tree**

  **Self Action Selection Tree**

[Demonstration format]
###
current_scene:
any_pot_not_full: false
any_soup_ready: true
holdings: {self: empty, teammate: dish}

inferred_teammate_current_skill: fill_dish_with_soup
self_action_now: pickup_dish

Explanation:
When a soup is ready, even if the teammate is already holding a dish and likely heading to fill it, self should adopt an aggressive strategy. Since the teammate's efficiency or final intent is uncertain, self should also pickup_dish to ensure that at least one person fills and delivers the soup quickly.
###
Scene <t>: <SELF_PLAYER> holds <item>. <TEAMMATE_PLAYER> holds <item>. Kitchen states: <pot/counter status>.
Analysis: <brief scene-level reasoning>
Plan for <SELF_PLAYER>: "<one legal action>".
###

[Memory and local reflection, appended after the first iteration]
<DECISION_TREE_MEMORY>
You are given some previous decision tree memories.
ALWAYS prefer decision trees with higher Final_Score, and reuse these to generate a new one.
Use the following memory for reference:
<POLICY_TREE_SUMMARY_MEMORY>

<THE_LAST_DECISION_TREE>
<LAST_ACCEPTED_POLICY_TREE>

<LAST_ISSUE_INFORMATION>
Analyze and modify the last decision tree using these improvement requirements:
<TREE_REFLEXION_JSON_WITH_DECISION_TREE_SUMMARY_REMOVED>

<OUTPUT_REQUIREMENTS>
- The output must start directly with "### FunctionDescription:" and end after the last line of the **Self Action Selection Tree**.
- Ensure that the generated decision tree covers the vast majority of branches in the game.
- Decide whether coordination is required. If coordination is required, condition self_action_now on inferred_teammate_current_skill in the **Partner Inference Tree**. If not, ignore inferred_teammate_current_skill.
- Provide one modified decision-tree function description strictly following the Required output format and <LAST_ISSUE_INFORMATION>. If no issue information is present, ignore it.
- Do NOT include explanations, introductions, or commentary outside the tree.
- Make sure the conditions do not conflict with each other.
- If any_soup_ready is true, pay attention to it.
\end{Verbatim}
\end{policybox}

\paragraph{Coder prompt.}
The coder receives the textual policy tree $T_k$ and converts it into an
executable Python function. The concrete template below instantiates the code
interface for Overcooked-AI action labels and state fields.

\begin{policybox}{Coder Prompt Template}
\begin{Verbatim}[fontsize=\tiny,breaklines=true,breakanywhere=true,breaksymbolleft={},breaksymbolright={}]
[System prompt]
Instructions:
You are an Overcooked_AI coder. Two players cooperate to maximize score by cooking onion soup (3 onions per pot), filling a dish, and delivering (+20). Pots auto-cook when full for 20 timesteps. Each player holds at most one item. No movement/location logic is needed here.

Each player may only choose a subset of these actions. Each action label must be written exactly in snake_case.
- <SELF_PLAYER> can execute: <SELF_SKILL_SET>
- <TEAMMATE_PLAYER> can execute: <TEAMMATE_SKILL_SET>

Your goal is to control <SELF_PLAYER> and cooperate with <TEAMMATE_PLAYER>.
- The players cannot communicate.
- Do not use move actions and do not use location information.
- You must do deliver_soup when you hold soup.
- If there is no ready soup in the current scene, you should not pickup_dish.
- You will be given a function description, a textual decision tree, in this format:
###
FunctionDescription:
  Name: PredictTeammateThenPlan
  Inputs:
    - current_scene:
        - holdings: {self: empty/onion/dish/soup, teammate: empty/onion/dish/soup}
        - pots: for each <Pot>: onion_count: {0,1,2,3}, state: {idle,cooking,ready}, timers
        - derived flags: any_soup_ready, any_pot_not_full
        - optional layout fields: num_empty_counters, num_onion_counters, num_dish_counters
    - teammate_last_completed_skill: one of the legal actions or None
    - teammate_last_inferred_skill: one of the legal actions or None
    - self_last_skill: one of the legal actions or None
  Outputs:
    - inferred_teammate_current_skill: one legal teammate action
    - self_action_now: one legal self action

  Partner Inference Tree:
    (textual if/else logic)

  Self Action Selection Tree:
    (textual if/else logic)
###

- Implement the decision tree as Python 3 using explicit if/elif logic.
- Implement tiny helpers if useful, e.g., _legalize(action, state_dict), to enforce hard legality rules and minimal fallbacks.
- Do not import external packages.
- Do not produce movement or location logic.
- Do not invent new action labels or change their spelling.

State input format:
###
state_dict = {
    'hold': [None, None],  # [player0_hold, player1_hold], each in {'empty','onion','dish','soup'}
    'pot': [],             # list of dicts: {'count': int, 'state': 'idle'|'cooking'|'ready', 'timers': int|None}
    'any_soup_ready': None,
    'any_pot_not_full': None,
    'teammate_last_completed_skill': None,
    'teammate_last_inferred_skill': None,
    'self_last_skill': None,
    # optional for Forced Coordination:
    'num_empty_counters': None,
    'num_onion_counters': None,
    'num_dish_counters': None,
}
###

Required output format:
```python
def PredictTeammateThenPlan(state):
    ...
    return inferred_teammate_current_skill, self_action_now
```

[Runtime user prompt]
Make sure to check whether the list variables are empty or not.
The tactic is:
<POLICY_TREE_T_k>

Please implement the code in Python format. Return the code wrapped with triple backticks (```python ... ```), and make sure the code is complete and syntactically correct.

[Repair prompt when generated code fails]
The current Python code implementation is:
<CURRENT_CODE>
When executing this code, the following error occurred:
<EXECUTION_ERROR>
Please carefully analyze the error message and modify the code to fix the issue.
\end{Verbatim}
\end{policybox}

\paragraph{Summarizer prompt.}
The summarizer receives the current policy tree and a language trace from the
executor, then returns a compact memory entry and a single local reflection for
the next planner call. The concrete template below instantiates the trajectory
diagnostics for Overcooked-AI rollouts.

\begin{policybox}{Summarizer Prompt Template}
\begin{Verbatim}[fontsize=\tiny,breaklines=true,breakanywhere=true,breaksymbolleft={},breaksymbolright={}]
[System prompt]
[Meta-Analysis Constraints]
You are part of an iterative policy refinement system with backtracking.

IMPORTANT:
- In each iteration, the decision tree is treated as FIXED except for ONE small local branch.
- Your role is NOT to evaluate the entire tree, but to help identify or justify a LOCAL adjustment.
- Even if the overall strategy seems reasonable, identify the most promising small inefficiency that could be improved.
- Avoid answering "no change" unless the episode shows near-perfect efficiency with no observable time waste.
- Prefer repeated throughput bottlenecks over isolated mistakes.
- Repeated illegal actions have the highest priority and should be selected before legal-but-low-value actions.
- Treat legal-but-low-value actions as important evidence if they repeatedly reduce soup throughput.
- Do NOT focus on teammate prediction errors unless they create repeated downstream delay or redundant work.
- Never propose an After_Behavior that violates action preconditions, such as selecting pickup_onion when self is holding a dish.

Assume:
- All existing rules, actions, and constraints are correct and must remain unchanged.
- Every proposed After_Behavior must respect the listed action preconditions.
- Reason about priorities, conditions, and coordination logic; do not invent new rules.

[Task and action knowledge]
Use the same Overcooked task rules, legal actions, role-specific action sets, and layout-specific symbolic fields as in the planner prompt.

Suppose you are an Overcooked_AI critic. You need to analyze decision quality. Do not propose or write code; do not rewrite the whole decision tree.

[Runtime prompt]
You are analyzing the behavior of a collaborative agent controlled by a FIXED decision tree, and then generating a Tree_Reflexion for a planner.

You MUST reason over the full episode trajectory across multiple scenes, instead of treating each scene independently.

IMPORTANT CONSTRAINTS:
- In the next iteration, ONLY ONE SMALL BRANCH of the decision tree can be modified.
- All other branches, priorities, and logic MUST remain unchanged.
- Do NOT propose a full rewrite.
- Do NOT suggest multiple alternative fixes.

Your goal is to identify the SINGLE MOST DAMAGING and REPEATABLE inefficiency that:
1) Causes clear throughput loss across multiple scenes, not just one isolated mistake,
2) Appears as repeated wasted timesteps, redundant role overlap, premature dish handling, delayed role switching, or legal-but-low-value actions,
3) Can still be attributed to ONE specific branch, priority rule, or missing condition in the decision tree.

PRIORITIZATION RULE:
- Prefer a bottleneck that repeatedly reduces soup throughput, even if all actions are technically legal.
- Do NOT over-prioritize teammate prediction mistakes unless they clearly create repeated downstream time loss.
- A valid-but-low-value action pattern is more important than a one-off invalid action.
- Also consider timing-related inefficiencies:
  - Acting too early or too late,
  - Situations where waiting would be better than acting immediately,
  - Situations where the agent chooses a legal action but the wrong role for the current phase.

[Episode states]:
Scene <t>:
<LANGUAGE_STATE_DESCRIPTION>
[Misprediction] <OPTIONAL_PARTNER_PREDICTION_ERROR>
[DecisionTrace] <OPTIONAL_SELECTED_SKILL_AND_EXECUTABILITY_TRACE>
###
... repeated for all logged scenes ...

[The decision tree]:
<POLICY_TREE_T_k>

[Final score]:
<S_k>

Your task:
1. Scan all episode states and identify time-wasting behaviors.
2. Select exactly ONE primary inefficiency with the largest negative impact on efficiency or score.
3. Identify the EXACT target branch or decision condition to modify.
4. Generate a Tree_Reflexion that guides a planner to perform a SINGLE, LOCAL modification:
   - describe the CURRENT behavior of this branch (BEFORE),
   - describe the DESIRED behavior after modification (AFTER), using refined conditions, priority changes, or a single added guard condition,
   - explicitly list assumptions about what logic MUST remain unchanged.
5. The value in Decision_Tree_Summary.Final_Score MUST be exactly <S_k>.

OUTPUT REQUIREMENTS:
- Output MUST be valid JSON.
- Do NOT include markdown, explanations, or extra text.
- Be concrete and implementable.
- Output must directly contain only these top-level keys:
  - "Decision_Tree_Summary"
  - "Tree_Reflexion"

Return JSON in the following format:
{
  "Decision_Tree_Summary": {
    "Summary": "Summarize the key strategies and critical decision points here.",
    "Final_Score": "<S_k>"
  },
  "Tree_Reflexion": {
    "Target_Branch": "",
    "Modification_Scope": "LOCAL_SINGLE_BRANCH",
    "Before_Behavior": "",
    "After_Behavior": [
      ""
    ],
    "Unchanged_Assumptions": [
      ""
    ],
    "Expected_Effect": ""
  }
}
\end{Verbatim}
\end{policybox}

\end{document}